\documentclass[12pt]{article}
\PassOptionsToPackage{numbers, compress}{natbib}
\usepackage[utf8]{inputenc}
\usepackage{amsmath}
\usepackage{amsfonts}
\usepackage{mathrsfs}
\usepackage{amssymb}
\usepackage{amsthm}
\usepackage{dsfont}
\usepackage{bm}
\usepackage{libertine}
\usepackage{wrapfig}
\usepackage{graphicx}
\usepackage{multirow}
\usepackage{epstopdf}
\usepackage{multicol}
\usepackage{subfigure}
\usepackage{booktabs}

\usepackage{algorithmic}
\usepackage{algorithm}
\usepackage[algo2e,linesnumbered]{algorithm2e}
\usepackage{stackengine}
\usepackage{graphicx}
\usepackage{multirow}
\usepackage{epstopdf}
\usepackage{multicol}
\usepackage{caption}
\usepackage{subfigure}
\usepackage{enumitem}
\usepackage{amsmath}
\usepackage{pifont}
\usepackage{natbib}
\usepackage{bbm}
\usepackage{makecell}
\usepackage{mathtools}
\usepackage[table]{xcolor}

\definecolor{mydarkred}{rgb}{0.6,0,0}
\definecolor{mydarkgreen}{rgb}{0,0.6,0}
\usepackage[colorlinks,
linkcolor=mydarkred,
citecolor=mydarkgreen]{hyperref}
\usepackage{url}
\usepackage{bm}
\usepackage{pifont}
\usepackage{stfloats}
\usepackage[T1]{fontenc}
\usepackage{arydshln}
\usepackage{colortbl}
\usepackage{balance}

\newtheorem{definition}{Definition}%

\newcolumntype{L}[1]{>{\raggedright\let\newline\\\arraybackslash\hspace{0pt}}m{#1}}
\newcolumntype{Y}{>{\centering\arraybackslash}X}
\newcolumntype{s}{>{\hsize=.3\hsize}Y}
\newcolumntype{t}{>{\hsize=1.5\hsize}X}
\newcolumntype{u}{>{\hsize=0.8\hsize}Y}
\usepackage{makecell}
\usepackage{mathtools}
\usepackage[utf8]{inputenc}
\usepackage{stfloats}

\title{Towards Harnessing Feature Embedding for \\ Robust Learning with Noisy Labels}

\author{
  Chuang Zhang\thanks{Chuang Zhang and Chen Gong are with PCA Lab, the Key Laboratory of Intelligent Perception and Systems for High-Dimensional Information of Ministry of Education, School of Computer Science and Engineering, Nanjing University of Science and Technology; Jiangsu Key Lab of Image and Video Understanding for Social Security. }
  \thanks{Part of this work was done during Chuang Zhang’s internship at JD Explore Academy.},~
  Li Shen\thanks{Li Shen is with JD Explore Academy.},~
  Jian Yang\thanks{Jian Yang is with College of Computer Science, Nankai University.},~
  Chen Gong\footnotemark[1]\thanks{Corresponding author.},\\
  \small{\{c.zhang,csjyang,chen.gong\}@njust.edu.cn;mathshenli@gmail.com}\\

}
\date{}
\begin{document}

\maketitle

\begin{abstract}
The memorization effect of deep neural networks (DNNs) plays a pivotal role in recent label noise learning methods. 
To exploit this effect, the model prediction-based methods have been widely adopted, which aim to exploit the outputs of DNNs in the early stage of learning to correct noisy labels.
However, we observe that the model will make mistakes during label prediction, resulting in unsatisfactory performance. By contrast, the produced features in the early stage of learning show better robustness. 
Inspired by this observation, in this paper, we propose a novel \emph{feature embedding-based} method for deep learning with label noise, termed \emph{\textbf{L}ab\textbf{E}l \textbf{N}oise \textbf{D}ilution} (LEND). To be specific, we first compute a similarity matrix based on current embedded features to capture the local structure of training data. Then, the noisy supervision signals carried by mislabeled data are overwhelmed by nearby correctly labeled ones (\textit{i.e.}, label noise dilution), of which the effectiveness is guaranteed by the inherent robustness of feature embedding. Finally, the training data with diluted labels are further used to train a robust classifier. Empirically, we conduct extensive experiments on both synthetic and real-world noisy datasets by comparing our LEND with several representative robust learning approaches. The results verify the effectiveness of our LEND.
\end{abstract}

\clearpage
\section{Introduction}\label{sec:introduction}
	The philosophy of recent success in deep learning mainly stems from massive high-quality labeled data, leading to impressive performance in countless areas, including computer vision \cite{krizhevsky2012imagenet, he2016deep}, natural language processing \cite{vaswani2017attention, devlin2018bert}, speech recognition \cite{hinton2012deep, sainath2013deep}, \textit{etc}.
	However, the data in real-world applications are often associated with label noise, due to human fatigue \cite{han2020survey}, knowledge limitation \cite{gong2017learning}, or measurement error ~\cite{song2020learning}. These noisy labels might degrade the performance of deep neural networks (DNNs) \cite{arpit2017closer, zhang2018generalized}, which raises a great demand for label noise-robust learning algorithms. Therefore, deep learning with label noise has been intensively studied due to its  wide applications \cite{li2017webvision,mahajan2018exploring,kuznetsova2018open}. 
	
	Previous results~\cite{arpit2017closer} suggest that DNNs first learn from examples with correct labels, and the noisy data will be fitted later. It is well-known as the memorization effect of deep learning.
	Inspired by this remarkable finding, a large group of previous works propose to employ the model predictions in the early stage of training to boost the robust learning.
	For model prediction-based robust methods, existing works can be generally attributed into two  categories, namely label correction-based methods and sample selection-based methods.
	Label correction-based methods \cite{li2020dividemix,xiao2015learning,yi2019probabilistic,tanaka2018joint,nguyen2019self} take the model predictions of training data as additional supervision signals to correct the potential  noisy labels for guiding DNNs' training. 
	Sample selection-based methods~\cite{jiang2018mentornet,han2018co,jiang2020beyond,han2020sigua,yao2020searching} select trustworthy examples with the loss value smaller than predefined thresholds during training to mitigate the negative effect of noisy labels.  
	
\begin{figure}[t]
    	\centering
    	\includegraphics[width=6in]{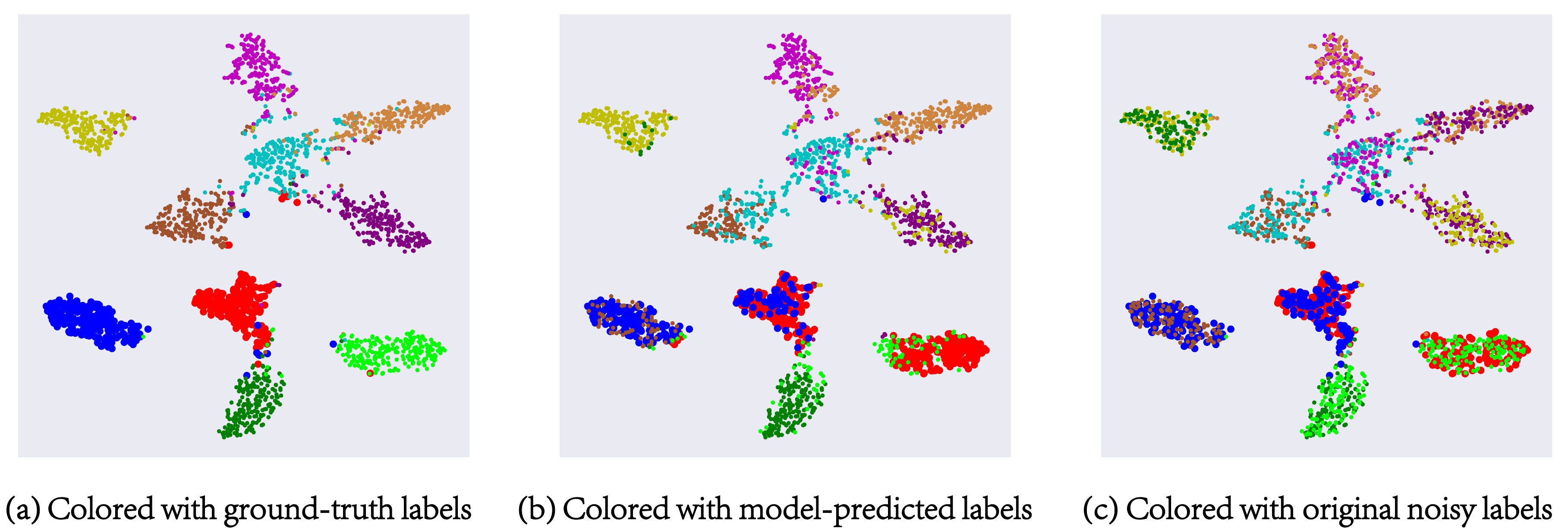}
    	\centering
    	\caption{The t-SNE visualization of training data under 30 epochs when training under asymmetric label noise (noise rate: 0.45) on \textit{CIFAR-10} dataset. ResNet-18 is directly used  to fit training data and the embedded features of the last hidden layer of the neural network are visualized by the t-SNE method. 
    	Subfigure (a) is colored with ground truth labels, (b) is colored with model-predicted labels, and (c) is colored with original noisy training labels.}
    	\label{fig:tsne}
	\end{figure}

    However, in practice, we observe that the model will make mistakes during label prediction and may be unstable in the output \cite{huang2020self}, 
    resulting in unsatisfactory performance. Fig.~\ref{fig:tsne} provides us a piece of evidence, from which we can see that the model still makes error-prone predictions even if the memorization effect exists, while the embedded features remain robust. 
    To be specific, we have the following three observations: 1) The model-predicted labels in Fig.~\ref{fig:tsne} (b) are much more reliable than the noisy ones in Fig.~\ref{fig:tsne} (c); 2) Under 30 epochs of training, the model still makes error-prone predictions for some training examples, especially for the cyan, red, and lime points in Fig.~\ref{fig:tsne} (b); 3) The embedded features in the early stage of learning contain strong semantic information, as the data points with the same ground-truth labels are clustered together in Fig.~\ref{fig:tsne} (a).
    It also indicates that the embedded features induced by the memorization effect are more robust than model predictions. The reason is that the classifier output following a neural network tends to fit the noise, while the embedded features are less negatively affected by the noise \cite{bai2021understanding}. Fig.~\ref{fig:acc} further provides us an empirical validation, where the accuracy of model-predicted labels measures the robustness of model predictions and the accuracy of diluted labels reveals the robustness of the feature embedding. While lots of previous works focus on developing robust models based on model predictions, scarce attention has been paid to the feature embedding. The above observations may provide us a new research insight for deep robust learning with noisy labels by leveraging the intrinsic feature embedding.

	In this paper, we claim that the embedded features induced by the memorization effect are more robust than the model-predicted labels. To instantiate such insight, we further propose a simple yet effective feature embedding-based label noise method, termed  \textbf{L}ab\textbf{E}l \textbf{N}oise \textbf{D}ilution 
    (LEND). 
    Therein, we first compute a similarity matrix based on current embedded features to capture the local structure of training data. Then, with the help of the trustworthy feature embedding, the noisy supervision signals carried by mislabeled data are overwhelmed by nearby correctly labeled ones, Finally, the corrected labels are further employed to train a robust classifier. 
    To be specific, for each example, its nearby examples first make voting in deciding whether its label is trustworthy, and then this message is propagated among its neighborhoods for further noise dilution. This process is conducted on the embedding space, of which the robustness has been validated before. Finally, the diluted labels are further employed to perform sample selection to train a robust classifier.
    Empirically, we conduct extensive experiments on both synthetic and real-world noisy datasets by comparing our method with several representative robust learning approaches, and the results verify the superiority of our LEND. 

	The contributions of this paper can be summarized as three-fold:
	\begin{itemize}
		\item [1).] We reveal a new finding that the embedded features induced by the memorization effect are more robust than the labels predicted by model, which provides us a new perspective for deep robust learning.
		\item [2).] We propose a novel feature embedding-based robust learning method, named LEND, which can make full use of the robust feature embedding in the early stage of learning.
		\item [3).] We evaluate our LEND on both synthetic and real-world noisy datasets, and the experiments well demonstrate its effectiveness.
	\end{itemize}
	
	The remaining of this paper is organized as follows: we first review the related works of two different categories of label noise learning methods in Sect. \ref{sec:related work}. In Sect. \ref{sec:problem setup}, we provide some preliminaries of deep label noise learning and detail the memorization effect of deep learning from the perspective of feature embedding. Further, we propose a simple yet effective label noise learning method in Sect. \ref{sec:method} and conduct extensive experiments on both synthetic and real-world noisy datasets to verify its effectiveness in Sect. \ref{sec:experiments}. Finally, we conclude our paper in Sect. \ref{sec:conclusion}.
\section{Related Work}\label{sec:related work}
    In this section, we provide a comprehensive review of the existing label noise learning studies, which can be roughly divided into two types, namely statistical learning-based methods and model prediction-based methods. 
    \subsection{Statistical Learning-Based Methods}
        This type of methods deal with the label noise learning problem by designing label noise-robust loss functions. 
        For example, \cite{natarajan2013learning} build an unbiased risk estimator to resist the adverse effect of the noisy labels. \cite{patrini2016loss} split the loss function into two parts in which only one of them is influenced by the corrupted labels. \cite{ghosh2017robust} provide some sufficient conditions on the loss function so that the risk minimization via using these loss functions would be inherently tolerant to label noise, such as Mean Absolute Error (MAE). Generalized Cross-Entropy (GCE) \cite{zhang2018generalized} can be interpreted as a generalization of MAE, as it integrates the advantages of both MAE and cross-entropy losses. Besides, DMI \cite{xu2019l_dmi} proposes an information-theoretic loss function, which utilizes Shannon’s mutual information and is robust to different kinds of label noise. 
        
        Another branch of works focus on modeling the label noise generating process by estimating the noise transition matrix and then designing a robust paradigm to correct the loss. 
        For example, \cite{goldberger2016training} explicitly model the label noise distribution by an additional softmax layer that connects the correct labels with the noisy ones.
        \cite{patrini2017making} correct the loss function by multiplying the noise transition matrix with the outputs of the softmax layer during forward propagation. However, the estimation of noise distribution often requires the existence of ``anchor points’’, which is an unpractical assumption in real-world applications.
        To address this drawback, 
        $T$-revision \cite{xia2019anchor} estimates the noise transition matrix without anchor points by adding fine-tuned slack variables. 
        \cite{li2021provably} simultaneously optimize two objectives, namely the cross-entropy loss between noisy labels and the model-predicted ones, and the transition matrix, which lead to an end-to-end framework for robust learning under label noise without anchor points.
        Besides, Class2simi \cite{wu2021class2simi} transforms data points with noisy class labels into data pairs with noisy similarity labels, so that the reduction of noise can be theoretically guaranteed. Moreover, \cite{yao2021instance} propose to model and make use of the causal process to correct noisy labels.

	\subsection{Model Prediction-Based Methods}
        This type of methods mainly combat noisy labels by exploiting the memorization effect. They aim to exploit the outputs of DNNs in the early learning stage and can be further divided into two categories, namely label correction-based methods and sample selection-based methods.
	
        \textbf{Label correction-based methods.} In the early learning stage, the model predictions are accurate on a subset of the mislabeled examples. This suggests that label correction-based methods are potential for correcting the corrupted labels during the robust training process. For example, \cite{tanaka2018joint} propose to adaptively correct noisy labels with new labels that are consistent with the probabilities estimated by the model. \cite{yi2019probabilistic} learn a set of extra hyperparameters to correct noisy examples. Usually, label correction is associated  with some iterative sample selection procedure or is with additional regularization terms. SELFIE \cite{song2019selfie} focuses on examples that have consistent model predictions, so as to minimize the falseness of label correction. \cite{DBLP:journals/corr/ReedLASER14} fits a two-component mixture model to perform sample selection, and then corrects labels via a convex combination. DivideMix \cite{li2020dividemix} uses two peer networks to perform sample selection via a two-component mixture model, and applies the semi-supervised learning technique MixMatch \cite{DBLP:conf/nips/BerthelotCGPOR19} to further correct the soft labels of the training data.

	    \textbf{Sample selection-based methods.} Sample selection-based methods try to pick up a relatively clean subset from the training set for model training.
	    For instance, MentorNet \cite{jiang2018mentornet} selects clean examples with the help of a pre-trained extra network, and then these selected examples are utilized to aid the training of the main network. Co-teaching \cite{han2018co} and Co-teaching+ \cite{yu2019does} adopt peer networks and use the ``small-loss'' and disagreement behaviors of the network to choose the possibly reliable examples to boost the robust learning of the network. Similarly, JoCoR \cite{wei2020combating} improves Co-teaching by utilizing co-regularization to reduce the diversity of the two networks. INCV \cite{chen2019understanding} utilizes cross-validation to filter out noisy examples at each training iteration. Besides, \cite{wang2018iterative} exploit the local outlier factor to select probable clean examples for training. \cite{nguyen2019self} introduce an extra mean teacher model to help the neural network to remove wrongly labeled data. S2E \cite{yao2020searching} proposes to employ an AutoML technique to dynamically decide the clean examples. 
        \cite{zhu2021second} propose to progressively sieve out corrupted examples with a robust peer loss, so that clean examples and the corrupted ones can be separately treated during training.
        Me-Momentum \cite{bai2021me} proposes to extract hard confident examples based on the memorization effect to obtain an accurate and stable decision boundary. 

\section{Basic Knowledge in Label Noise Learning}\label{sec:problem setup}
    In this section, we first provide some preliminaries of deep label noise learning and then discuss the memorization effect from the perspective of feature embedding.
    
	\subsection{Preliminaries}
	In a traditional supervised $C$-class classification problem, the training data $\mathcal{D}=\{(x_i,y_i)\}_{i=1}^n$ of size $n$ are randomly sampled from the joint distribution $p(X,Y)$, where $x$ refers to the feature of example, $y$ denotes the corresponding true label, and $X$ and $Y$ denote the corresponding random variables. The training data are used to fit a DNN $f(x;\theta):\mathcal{X}\rightarrow\mathbb{R}^C$ parameterized with $\theta$ that conceptually consists of two parts. The first part is a representation function $\phi(x):\mathcal{X}\rightarrow \mathbb{R}^d$ that maps the input $x$ to a $d$ dimensional embedding space, and the second one is a classifier following $\phi(x)$ with a soft-max link function, generating the confidence scores of $x$ regarding every category. 
	
	When the labels are corrupted with external noise, the observed noisy labels $\tilde{\mathcal{Y}}=\{\tilde{y}_i\}_{i=1}^n$ are  used during training. It means that the training data are drawn from $p(X,\tilde{Y})=\sum_{Y} p(\tilde{Y}\mid  Y,X)p(Y\mid  X)p(X)$, where $p(\tilde{Y}\mid  Y,X)$, termed as the label transition probability, characterizes label corruption. Here are three common specifications of $p(\tilde{Y}\mid  Y,X)$:
	\begin{itemize}
		\item[1)] \textit{Symmetric label noise}: with the uniform distribution of noisy labels, \textit{i.e.,} $p(\tilde{Y}=\tilde{y}\mid  Y=y,X)=\rho/(C-1),\forall~\tilde{y}\ne y$ and $\rho$ is the noise rate, noisy labels are produced at random;
		\item[2)] \textit{Asymmetric label noise}: with the label-dependent probability $p(\tilde{Y}\mid  Y,X)=p(\tilde{Y}\mid  Y)$, some pairs of classes are more prone to be mislabeled~\cite{patrini2017making};
		\item[3)] \textit{Feature-dependent label noise}: with the full feature-dependent probability $p(\tilde{Y}\mid  Y,X)$, the label noise follows a more realistic label noise distribution~\cite{xiao2015learning,jiang2020beyond}. 
	\end{itemize}
	Given the noisily labeled training data, robust learning methods are required, such that the resulting DNNs can predict true labels. In the literature, existing works can generally be attributed to two categories: 1) statistical learning-based methods and 2) model prediction-based methods.
    Even with rigorous theoretical guarantees, the former may not work well for DNNs~\cite{han2020survey}. In contrast, the latter is designed specifically for deep learning with label noise~\cite{jiang2018mentornet}. 
    Our work is different from any of the above two kinds of methods. Inspired by the robustness of the feature embedding in the early stage of learning, we focus on designing a new robust learning method under label noise, termed \textit{feature embedding-based} method.

\subsection{Rethinking the Memorization Effect}\label{sec:rethinking}

The memorization effect in deep learning with noisy labels suggests that DNNs will first fit correctly labeled examples and learn from the noisy ones later. In view of optimization, it means that the correctly labeled examples dominate the updating directions of stochastic gradient descent early in DNNs' training. In previous works, the memorization effect of DNNs is mainly studied with respect to class predictions by model. Here, the memorization effect is further explored from the perspective of representation learning, \textit{i.e.}, feature embedding.
We claim that the memorization effect induces a good feature embedding, and the induced embedded features are more robust to label noise than the corresponding model predictions. In the following, we provide a piece of empirical evidence and a more detailed discussion.

 Fig.~\ref{fig:tsne} provides the t-SNE visualization of the features of training data under 30 epochs when training under asymmetric label noise on \textit{CIFAR-10} dataset.
We can see that 1)  model-predicted labels are much more reliable than the original noisy ones; 2) In the early stage of learning,  the model still makes mistakes during prediction; 3)  The data points with the same ground-truth labels tend to fall into the same cluster.
 The first observation reflects that the DNNs tend to fit the examples with correct labels in the early learning stage, and the model predictions are able to recover the
distribution of the ground-truth labels to some extent. This observation confirms the conclusion about the memorization effect in \cite{arpit2017closer}.
 The second observation reveals the deficiency of model-predicted labels, \textit{i.e.}, the classifier still fits some noise even if the memorization effect exists.
 As for the third observation, it indicates that the noisy labels have relatively little effect on the embedded features compared with the model-predicted labels. 
 These observations provide us a new thinking to memorization effect, \textit{i.e.}, the embedded features in the early stage of learning are more robust than model predictions.
 
 To further validate the observations aforementioned, we provide an empirical validation in Fig.~\ref{fig:acc}. The accuracy of model-predicted labels measures the robustness of model predictions and the accuracy of diluted labels reveals the robustness of feature embedding. Besides, more visualizations of the training process with noisy labels are provided in Appendix~\ref{sec:tsne}. 
Concretely, we provide the t-SNE visualizations on \textit{CIFAR-10} dataset with asymmetric label noise (noise rate: 0.45) when the model is trained under 50, 100, 150, and 200 epochs. 
We can see that in the early stage of learning, both feature embedding and model predictions inherit useful information from the memorization effect, and feature embedding shows better robustness than model-predicted labels (see Fig. \ref{fig:tsne50} and \ref{fig:tsne100}). When trained to converge (see Fig. \ref{fig:tsne200}), DNNs will often fit and memorize noisy labels \cite{arpit2017closer},  and thus the data points belonging to the same cluster (\textit{i.e.}, have the same ground-truth label) will often be assigned with different labels.

\section{Our Method}
	\begin{wrapfigure}{r}{0.52\textwidth}
	    \vspace{-0.6cm}
		\centering
		\includegraphics[width=2in]{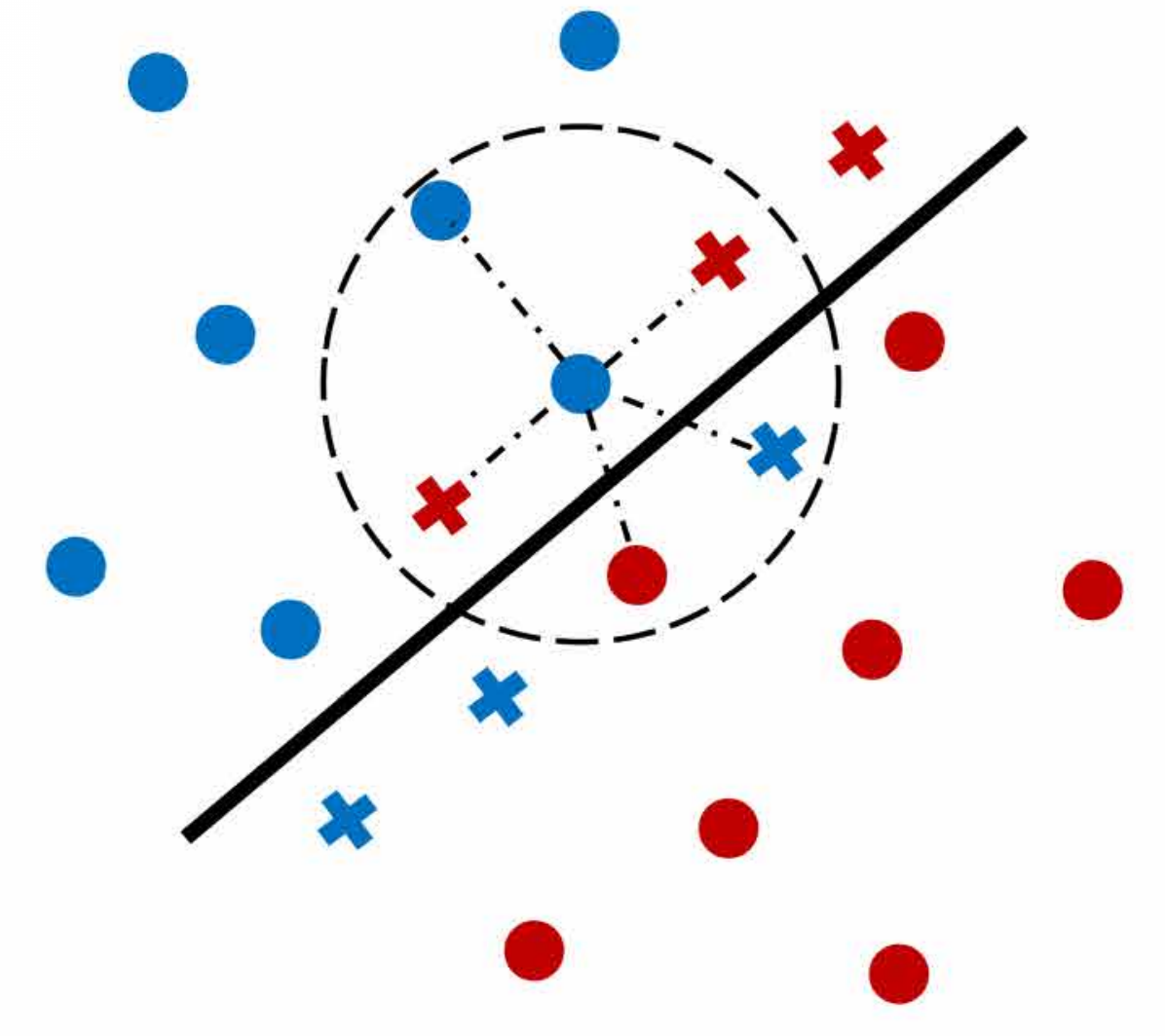}\\
		\caption{Examples that are close to the potential decision boundary, where the examples with different labels are represented by different colors, and the shape denotes whether the example is correctly labeled (``$\times$'' for corrupted and ``$\bullet$''for clean).}
		\vspace{-1.8cm}
		\label{fig:kNN}
	\end{wrapfigure}
        
	\label{sec:method}
    In this section, we present our simple yet effective feature embedding-based label noise-robust method termed ``LEND''. We first discuss how to employ the robustness of feature embedding and then detail the proposed LEND.

	\subsection{Feature Embedding-Based Label Noise Learning}
 	
	First, we conduct an analysis on feature embedding for the further employment. 
    To this end, the definition of \textit{dominant label} is provided, which is based on the feature embedding $\phi(x)$, namely:

	\begin{definition}[Dominant Label] \label{def:hard example}
		\textit{The dominant label $m_i$ of an example $x_i$ is defined as:}
		\begin{equation*}
			m_i \leftarrow \arg\max_{c} \left\{\sum{\mathds{1}(y_j=c)} \mid j\in \text{NN}_{k}(x_i;\phi)~\wedge~c \in \{1,\cdots,C\}\right\},
		\end{equation*}
		\textit{where $\mathds{1}(\cdot)$ is an indicator function which equals to one if its argument is true, and zero otherwise, $\text{NN}_{k}(x_i;\phi)$ includes the indices of examples which are the $k$ nearest neighbors of $x_i$ in the training set in terms of the feature embedding $\phi(x)$.}
	\end{definition}
	
	The dominant label of an example $x_i$ is the label that occurs most frequently in $x_i$'s $k$ nearest neighbors, which can be viewed as a notion of agreement on the local data information regarding the feature embedding $\phi(x)$.
    Intuitively, if the embedded features retain strong semantic information regarding ground-truth labels, the neighborhoods of each example $x_i$ are very likely to share the same ground-truth label.

	Therefore, in this section, we aim at devising a label noise learning method based on the previous finding. 
	One na\"ive solution is to directly use the dominant labels after every iteration as strong supervision signals. Unfortunately, although the ability of $k$NN-based methods in filtering label noise has been verified \cite{bahri2020deep}, it implicitly assumes that, for most training examples, the majority of their neighborhoods are correctly labeled. In practice, this assumption may fail for the examples that are close to the potential decision boundary. These examples are located in a region with complex semantic context, where the employment of dominant labels may fail. For example, when an example is close to the decision boundary (see Fig. \ref{fig:kNN}), its $k$ ($k=5$) nearest neighbors may include examples that cross over the decision boundary, and the dominant label-based method will not work.
	In the following parts, we devise a simple yet effective label noise-robust method to overcome such difficulty.

	\subsection{Label Noise Dilution}
\begin{figure*}[t]
		\centering
		\includegraphics[width=6in]{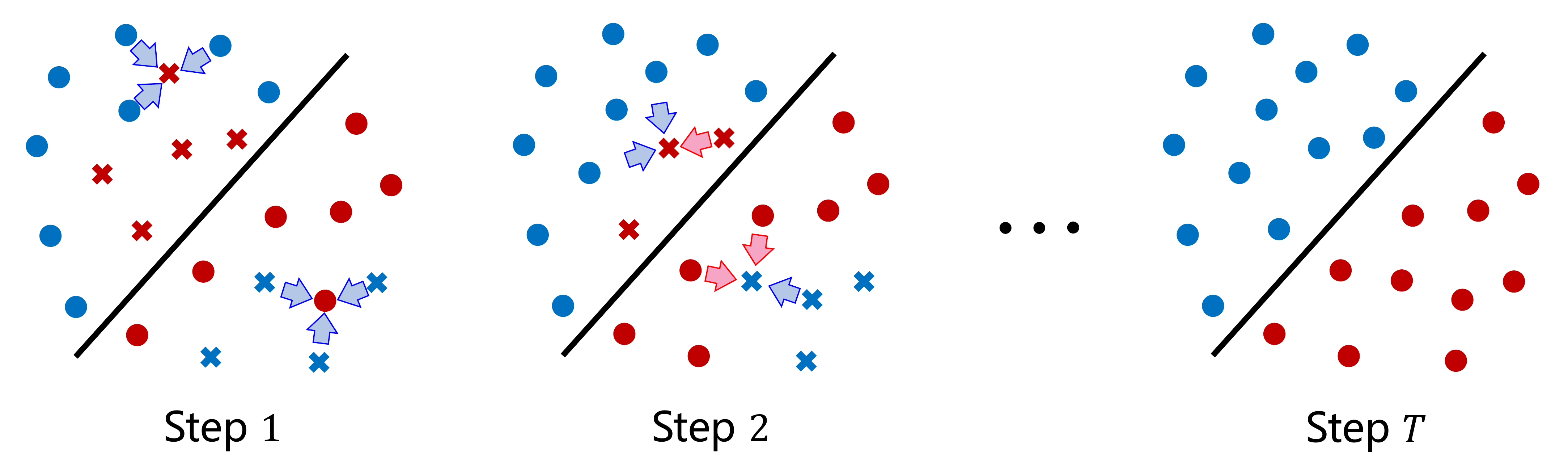}
		\caption{Illustration of label noise dilution. The leftmost figure represents the original data with noisy labels. 
		Therein, the examples with different labels are represented by different colors, and the shape denotes whether the example is correctly labeled (``$\times$'' for corrupted and ``$\bullet$''for clean).
		We can see that the examples with corrupted labels are gradually revised by the nearby correctly labeled ones. After several iterations of dilution, the corrupted labels are removed.}
		\label{fig:insight_dilution}
		\vspace{-0.2cm}
	\end{figure*}

	To overcome the dilemma that some examples may fall into the region with complex semantic context, in this section, we devise a robust label noise dilution paradigm. Therein, definite class information can be diffused to the region with ambiguous class information. As a result, the corrupted labels of training data are ``diluted'' by strong semantic information from correctly labeled ones. Its high-level idea is that the propagated information from correctly labeled examples will gradually cleanse the inaccurate  labels. This attempt is mainly inspired by label propagation in semi-supervised learning~\cite{iscen2019label}. Fig.~\ref{fig:insight_dilution} presents our main motivation.

    As analyzed before, the feature embedding conveys strong semantic information regarding ground-truth labels. 
    Given training data $\mathcal{B}=\{(x_i,\tilde{y}_i)\}_{i=1}^{\lvert \mathcal{B} \rvert}$ with $\lvert \mathcal{B} \rvert$ being the amount of input examples in the current mini-batch, we aim to dilute the label noise by using the embedded features $\{\phi(x_i)\}_{i=1}^{\lvert \mathcal{B} \rvert}$. 
    
    First, we specify a similarity matrix $W \in \mathbb{R}^{\lvert \mathcal{B} \rvert \times \lvert \mathcal{B} \rvert}$ which captures the local information of each data point of training data. Following \cite{iscen2019label}, given the embedded features $V \in \mathbb{R}^{\lvert \mathcal{B} \rvert \times d}$
    with $v_i={\phi}(x_i)$, we first construct the sparse similarity matrix $A\in \mathbb{R}^{\lvert \mathcal{B} \rvert \times \lvert \mathcal{B} \rvert}$ with elements:
	\begin{equation}
		A_{ij} = \left\{
		\begin{aligned}
			& [v_i v_j^\top]_{+}^\gamma, & \mathrm{if} ~ i\ne j ~\text{and}~ j\in \text{NN}_k(x_i;\phi), \\
			& ~~~0, & \text{otherwise},~~~~~~~~~~~~~~~\label{eq:affine}
		\end{aligned}
		\right.
	\end{equation}
	where $\gamma$ is a scaling parameter by following \cite{iscen2017efficient} and $[v_i v_j^{\top}]_{+}=\operatorname{max}(0,v_i v_j^\top)$.
	Following that, we have the final similarity matrix $W=D^{-1/2}W' D^{-1/2}$ with $W'=A^\top A$ 
	and  $D=\text{diag}(W' \boldsymbol{1})$ being the degree matrix, where $\boldsymbol{1}$ is an all-one column vector.
	
    Then, we detail the label noise dilution process by using the  similarity matrix $W$. In each iteration of label noise dilution, for every example $x_i$, we first get its $k$ nearest neighbors according to the representation $\phi(x)$, and then we integrate  the label information from the $k$ nearest neighbors with its own label in the previous iteration. 	 
	Formally speaking, let $Z \in \mathbb{R}^{\mathcal{\lvert \mathcal{B} \rvert \times C}}$ denote the one-hot encoded diluted labels, we update the diluted label $Z_i$ (the $i$-th row of $Z$) of example $x_i$ in the $t$-th iteration as:
	\begin{equation}
		Z_i^{(t)} = \alpha \sum_{j\in \text{NN}_k(x_i;\phi)} W_{ij} Z_j^{(t-1)} + (1-\alpha) Z_i^{(t-1)},
		\label{eq:z}
	\end{equation}
	where $Z_i^{(t)}$ denotes the diluted label of example $x_i$ in the $t$-th iteration, the first term of the right-hand side of Eq.~\eqref{eq:z} represents the label information from $x_i$'s $k$ nearest neighbors, and the second term denotes the reserved label information of $x_i$ from the $(t-1)$-th iteration. Here, $W_{ij}$ characterizes the similarity between $x_i$ and its $j$-th nearest neighbor,  and $0<\alpha<1$ is a trade-off parameter. The diluted labels of the training data  can be concisely written in matrix form as:
	
	\begin{equation}
		Z^{(t)} = \alpha  W Z^{(t-1)} + (1-\alpha)  Z^{(t-1)} .
		\label{eq:label_prop}
	\end{equation}
	At the beginning, the diluted labels $Z^{(0)}$ of training examples are initialized by one-hot representation of the corresponding noisy labels, which are denoted as: 
	\begin{equation}
	   Z_{i}^{(0)} := \texttt{onehot}(\tilde{y}_i),~~~~~~~~i=1,\ldots,\lvert \mathcal{B} \rvert.
	   \label{z0}
	\end{equation}
	 Then, we use Eq.~\eqref{eq:label_prop} to update the diluted labels of each example until converge. 
    
\begin{wrapfigure}{r}{0.5\textwidth}
		\centering
		\vspace{-0.1cm}
		\includegraphics[width=2.5in]{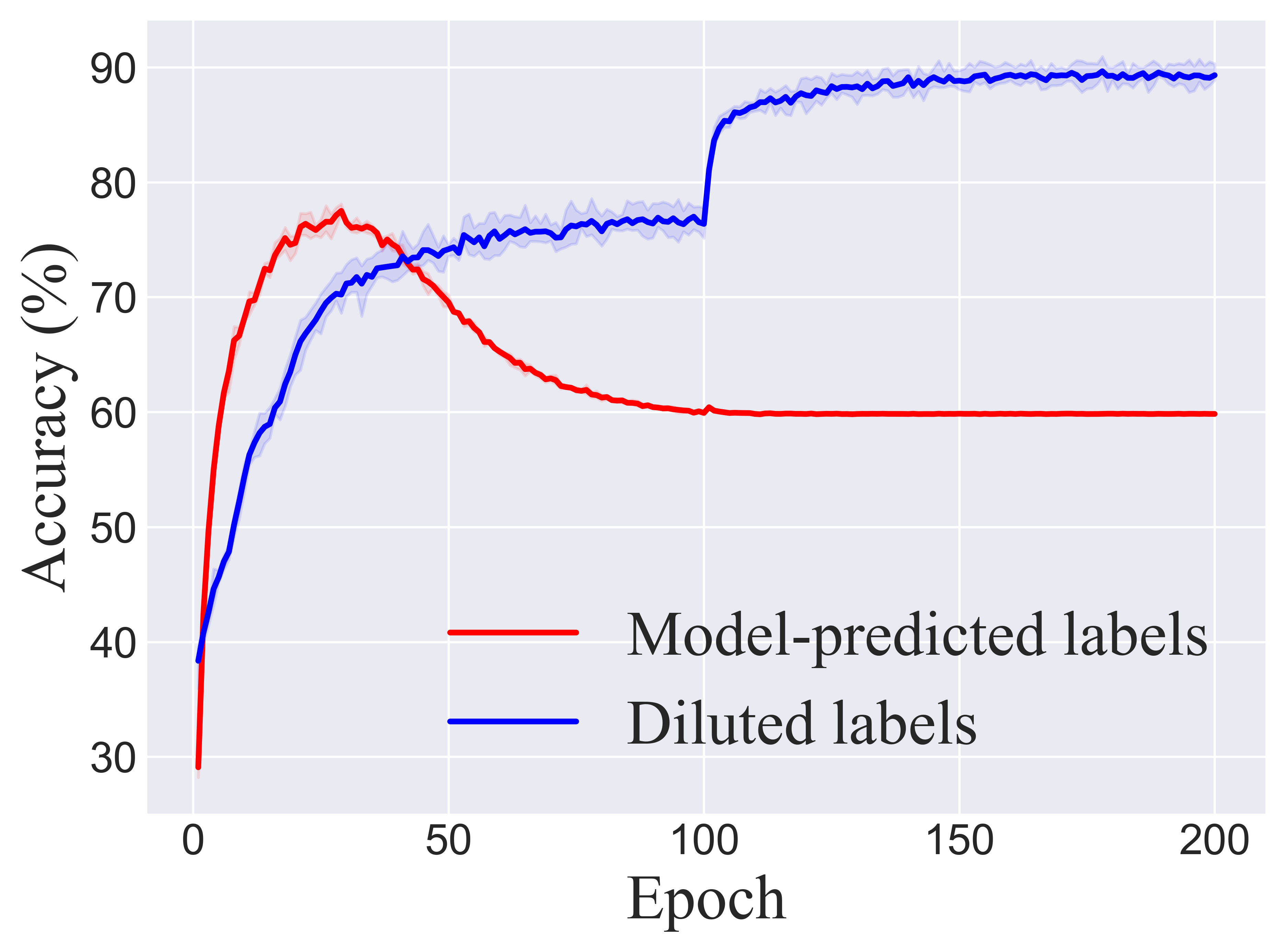}
		\caption{Accuracy curves with respect to model-predicted labels and diluted labels, where \textit{CIFAR-10} dataset under $40\%$ asymmetric label noise is considered. }
		\label{fig:acc}
	\vspace{-0.7cm}
	\end{wrapfigure}

    Here, we conduct a quantitative analysis to validate the effectiveness of the label noise dilution process. Specifically, we compare the robustness of DNNs with respect to model predictions and embedded features. The former is measured via the accuracy of model-predicted labels, and the latter is revealed by the accuracy of diluted labels. 
    Fig. ~\ref{fig:acc} shows the accuracy curves of diluted labels and model-predicted labels during training. We can see that the accuracy of diluted labels outperforms that of model-predicted ones with a large margin, validating the robustness of the feature embedding. In the following part, we further provide how to use the diluted labels $Z$ to train label noise-robust DNNs.

\subsection{Feature Embedding-Based Sample Selection}

\renewcommand{\algorithmicrequire}{ \textbf{Input:}}
\renewcommand{\algorithmicensure}{ \textbf{Output:}}
\begin{algorithm}[t]
\small
	\caption{The overall algorithm of LEND.} 
	\begin{algorithmic}[1]
		\REQUIRE Training dataset $\mathcal{D} = \{(x_i,\tilde{y}_i)\}_{i=1}^{n}$, initial parameters $\theta$ of DNN, momentum $\beta$, number of nearest neighbors $k$; 
		\WHILE{$\tau<\text{MaxEpoch}$}
			\FOR{$iter=1$ \textbf{to} $num\_iters$}
				\STATE Get current model $f(x;\theta)$ with underlying representation function $\phi(x)$;
				\STATE From $\mathcal{D}$ draw a mini-batch of data $\mathcal{B}=\{(x_i,\tilde{y}_i)\}^{\lvert \mathcal{B} \rvert}_{i=1}$ with batch-size $\lvert \mathcal{B} \rvert$;
				\STATE \emph{// \color{blue}Calculate the similarity matrix}
				\FOR{$j, k =  {1,\ldots, \lvert \mathcal{B} \rvert}$}
					\STATE $A_{j,k}\leftarrow$ similarity values via Eq.~\eqref{eq:affine};
				\ENDFOR
				\STATE $W'= A^\top A$ and $D=\text{diag}(W' \boldsymbol{1})$;
				\STATE $W=D^{-1/2}W' D^{-1/2}$;
				\STATE \emph{// \color{blue}Label noise dilution}
				\STATE \textbf{Initialize} the diluted labels of current mini-batch $Z^{(0)}$ via Eq. \eqref{z0};
				\FOR{$t\in \{1,\cdots, T\}$}
					\STATE Compute the diluted labels of current mini-batch $Z$ via Equation \eqref{eq:label_prop};
				\ENDFOR
				\STATE Update the diluted labels at current epoch $Z^{\tau}$ via Eq. \eqref{momentum}; 
				\STATE \emph{// \color{blue}Sample selection with diluted labels}
				\STATE Compute the discrete sample weights $E$ via Eq. \eqref{weights};
				\STATE  Update the model parameters $\theta$ via Eq. \eqref{sgd};
				\STATE Update $\tau \leftarrow \tau +1$;
			\ENDFOR
		\ENDWHILE
		\ENSURE The updated parameters $\theta$.
	\end{algorithmic}
	\label{alg}
\end{algorithm}

In this part, we present 1) how to utilize the label noise dilution in stochastic gradient-based optimization; and 2) how to use the diluted labels to train a robust DNN.
    
For the first problem, at the $\tau$-th epoch of  training, given a mini-batch of  training data $\mathcal{B} = \{(x_i, \tilde{y}_i)\}_{i=1}^{\lvert \mathcal{B} \rvert}$, we first compute a similarity matrix $W$ according to the current representation $\phi(x)$ of the examples $x$ within this mini-batch. Then, we can get the corresponding diluted labels $Z$ via Eq.~\eqref{eq:label_prop}.  
To make full use of the memorization effect in the early learning stage, we further employ a running average for the diluted labels. That is to say, the diluted labels of mini-batch $\mathcal{B}$ at the $\tau$-th epoch of training, \textit{i.e.}, $Z^{\tau}$, can be updated as:
\begin{equation}
	Z^{\tau} = (1-\beta) Z + \beta Z^{\tau-1}.
	\label{momentum}
\end{equation}
Therein, $\beta$ is the momentum  and $Z^{\tau-1}$ is the diluted labels of training data in $\mathcal{B}$ from the $(\tau-1)$-th epoch of training. 

    Note that the robustness of diluted labels has been verified in Section \ref{sec:introduction}. For the second problem, we employ the diluted labels $Z^{\tau}$ to aid sample selection and the resulting confident clean examples are used to update the parameters of DNN. To be specific, let $E = \{e_i\}^{\lvert \mathcal{B} \rvert}_{i=1}$ be the $\{0,1\}$-binary sample weights, which represent whether the training examples are selected, we can get $e_i$ via the following equation:
    \begin{equation}
    	e_i=\mathds{1}(\tilde{y}_i=\arg\max_j Z^{\tau}_{i,j}).
    	\label{weights}
    \end{equation}
        Note that the diluted labels $Z^{\tau}$ are encoded with one-hot representation, and $Z^{\tau}_{i,j}$ denotes the $j$-th value of the $i$-th example of $\mathcal{B}$ at the $\tau$-th epoch of training.
        With the sample weights, the parameters $\theta$ of DNN can be updated by stochastic gradient descent as:
        \begin{equation}
        \theta\leftarrow \theta-\nabla_{\theta} \sum_{(x_i,\tilde{y}_i)\in \mathcal{B}} e_i \ell(f(x_i,\theta),\tilde{y}_i),
        \label{sgd}
        \end{equation}
        where $\ell(\cdot)$ denotes a specified loss function during DNN's training and $\nabla_{\theta}$ represents the gradient with respect to $\theta$. The overall algorithm is summarized in Algorithm \ref{alg}.

\section{Experiments}\label{sec:experiments}

	This section examines the robustness of the proposed LEND by comparing it with several existing representative label noise learning methods.	Therein, extensive experiments are conducted on commonly used image classification datasets, namely, \textit{CIFAR-10}, \textit{CIFAR-100}, and \textit{Animal-10N}. For \textit{CIFAR-10} and \textit{CIFAR-100}, synthetic label noises with various noise types and levels are manually added. \textit{Animal-10N} is a real-world noisy dataset, which is used to testify the effectiveness of our method in dealing with practical label noise.

	\subsection{Datasets}

	\textit{CIFAR-10} and \textit{CIFAR-100} \cite{krizhevsky2009learning} are used to verify the efficacy of our approach in dealing with symmetric and asymmetric label noise, which are popularly used for evaluating the learning methods with noisy labels \cite{han2018co,yu2019does,jiang2018mentornet}. 
	Specifically, both datasets consist of 50,000 images for training and 10,000 images for testing, and the image size is 32$\times$32. The data in \textit{CIFAR-10} are classified into 10 categories, while \textit{CIFAR-100} considers a 100-category classification problem. Since all original datasets are clean, following the common setting in \cite{patrini2017making}, we utilize two representative structures of noise transition matrix to convert the original clean labels to noisy ones, which are

	\begin{itemize}
		\item \textbf{Symmetric flipping}~\cite{van2015learning}: the labels are randomly flipped to all possible categories with a certain probability. 
		\item \textbf{Asymmetric flipping} \cite{han2018co}: the labels are randomly flipped into  similar classes (\textit{e.g.}, $\text{DEER}\leftrightarrow \text{HORSE}$, $\text{DOG} \leftrightarrow \text{CAT}$). 
	\end{itemize}
	
	\begin{wrapfigure}{r}{0.5\textwidth}
	 \vspace{-0.55cm}
		\centering
		\includegraphics[width=3 in]{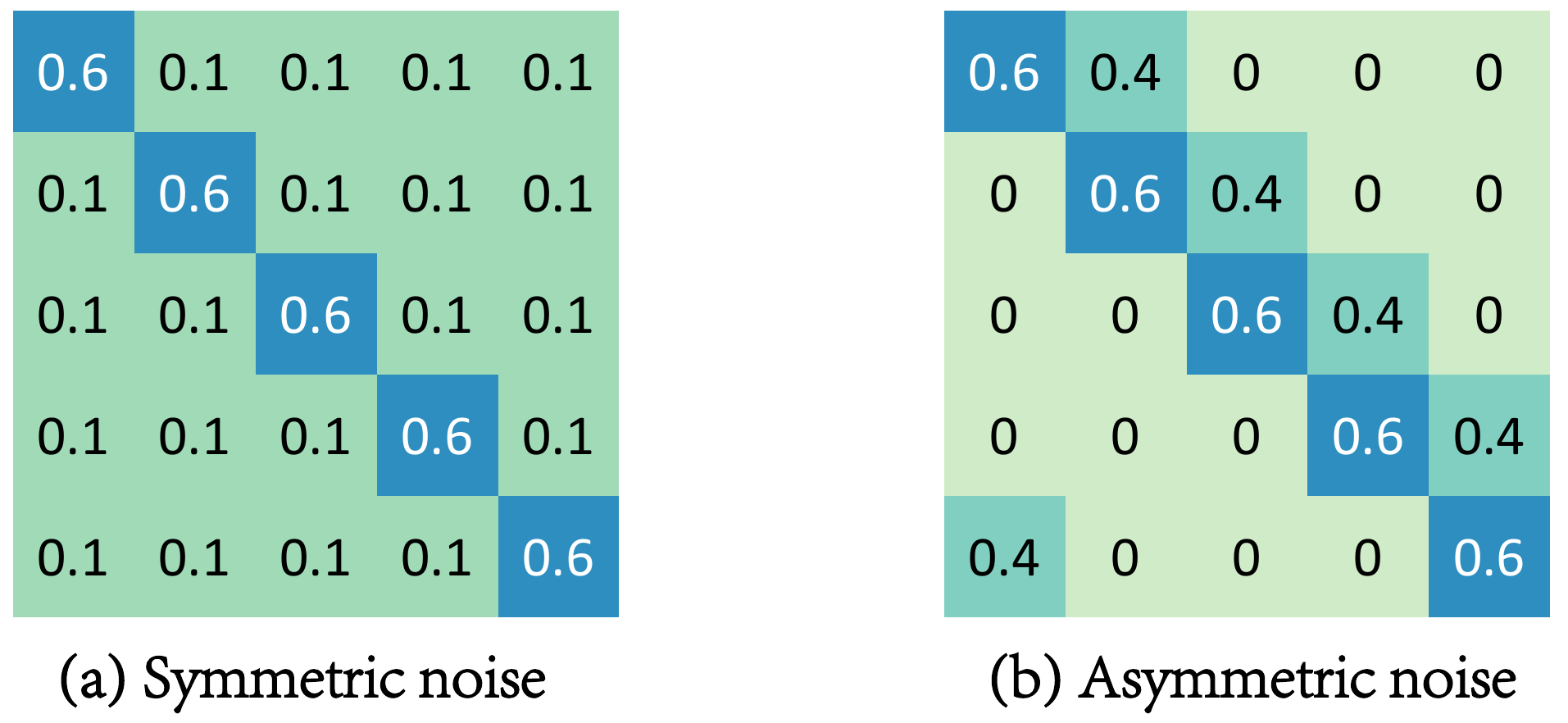}\\
		\caption{Examples of noise transition matrix of (a) symmetric noise and (b) asymmetric noise, 
		where there are five classes in total and the noise rate is 0.4.}
		\label{fig:noise_type}
		\vspace{-0.4cm}
	\end{wrapfigure}
	
	\textit{Animal-10N} is a real-world noisy dataset with natural label noise. It is introduced by \cite{song2019selfie} first and is constituted by five pairs of confusing animals. The images are crawled from several online search engines including \textit{Bing} and \textit{Google} using the predefined labels as the search keywords. All label noise on \textit{Animal-10N} is introduced by human mistakes, and the overall noise rate is around $8\%$. In total, \textit{Animal-10N} contains $50,000$ RGB images used for training and $5,000$ RGB images for testing, and the resolution of each image is $64\times64$.
	
	\subsection{Compared Baseline Methods}
	We compare the proposed method with the following representative label noise-robust learning algorithms: 
	
	\begin{itemize}
		\item  \textbf{Standard}, a simple baseline method, where DNNs with cross-entropy loss are directly trained on noisy datasets. It can be viewed as a competitor without tackling label noise.
		\item  \textbf{Co-teaching} \cite{han2018co}, a representative sample selection-based method, which trains two networks in a collaborative way by back-propagating the peer error.
		\item  \textbf{GCE} \cite{zhang2018generalized}, a loss correction-based method where the generalized cross-entropy loss is used for training a robust neural network.
		\item \textbf{JoCoR} \cite{wei2020combating}, a sample selection-based method that trains two networks and utilizes co-regularization to reduce the diversity of the two networks.
		\item \textbf{SIGUA} \cite{han2020sigua}, a gradient correction-based method, which adopts gradient descent on clean data while applying a learning-rate-reduced gradient ascent on noisy data.
	\end{itemize}
	
	Notably, for a fair comparison, the backbone network architectures are the same for our LEND and the compared baseline methods, and all methods are implemented by PyTorch.

	\subsection{Experiments on Synthetic Dataset}

	\paragraph{\textbf{Implementation Details}}
	For a fair comparison, all experiments are conducted on one Titan-V GPU. We employ ResNet-18 as the backbone network for all competitors. For our LEND, the model is trained under 200 epochs, and we adopt SGD with a momentum of 0.9, a weight decay of 0.0005, and a batch size of 256. 
	The initial learning rate is set to 0.05, and divided by 10 after 100 epochs. Besides, if not specified, the trade-off parameter $\alpha$ in Eq.~\eqref{eq:z} is set to 0.99 by following \cite{iscen2017efficient} and the momentum $\beta$ in Eq. \eqref{momentum} is set to 0.9. We provide a parameter sensitivity analysis regarding $\beta$ in Section \ref{para_ana}.
	Fig.~\ref{fig:noise_type} provides us the examples of the noise transition matrix for both symmetric and asymmetric noise, where a five-class classification problem with the noise rate $\rho=0.4$ is taken as an example.

	    \begin{figure*}[t]
	\begin{minipage}{1\linewidth}
		\centering
		\subfigure[Symmetric-40\%.]
		{\includegraphics[width=0.3\linewidth]{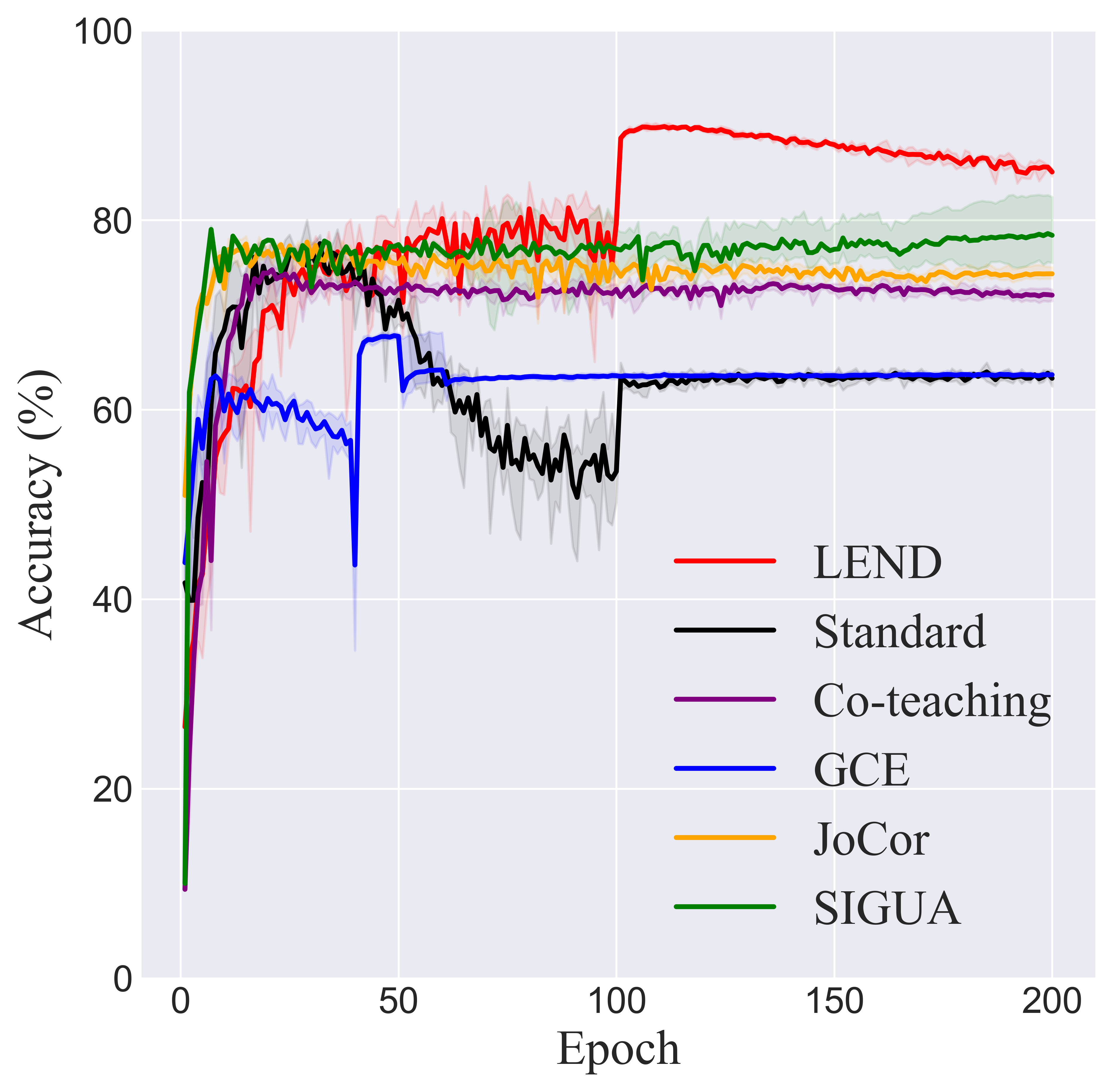}}		
		\subfigure[Symmetric-50\%.]
		{\includegraphics[width=0.3\linewidth]{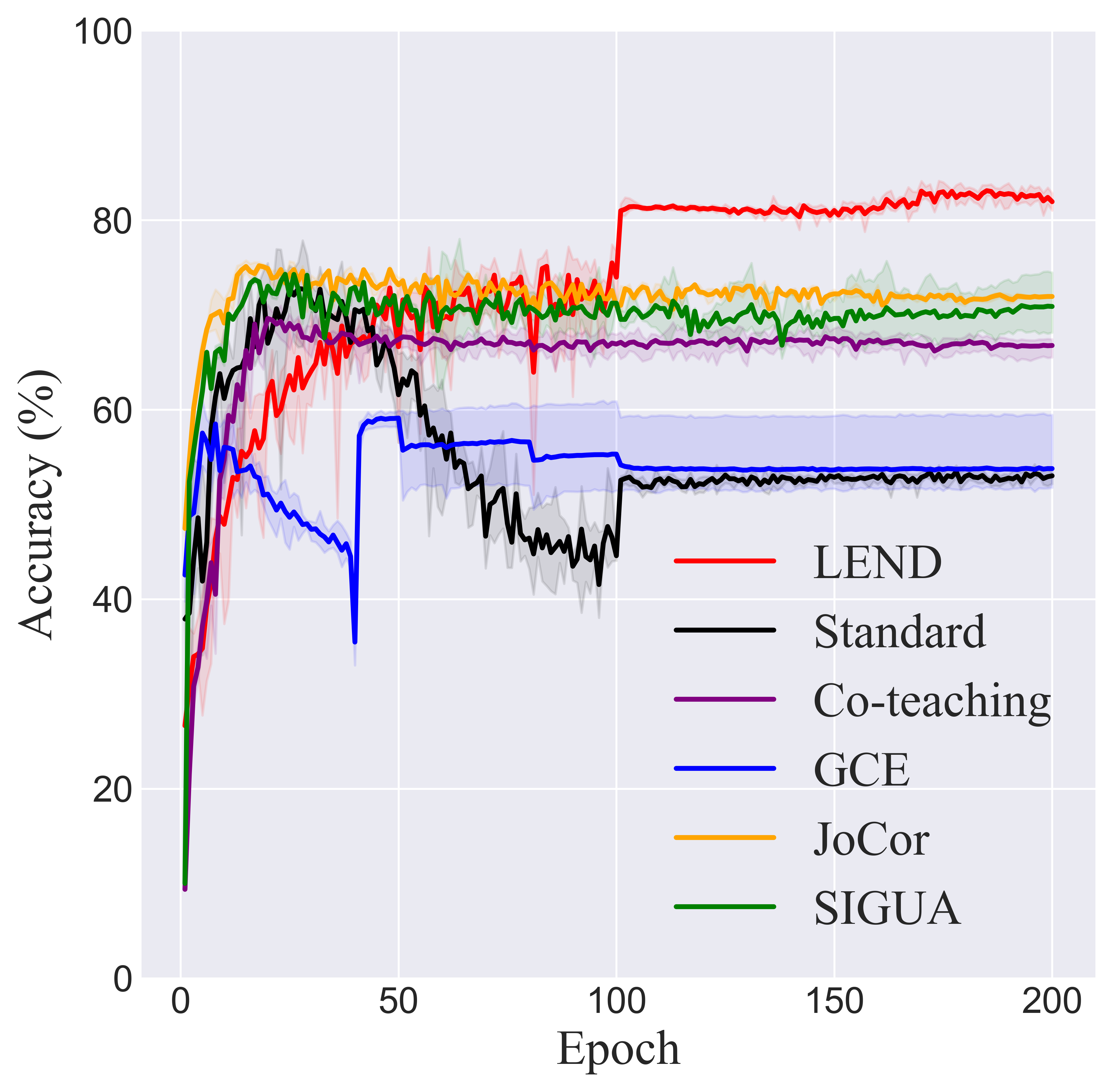}}
		\subfigure[Symmetric-60\%.]
		{\includegraphics[width=0.3\linewidth]{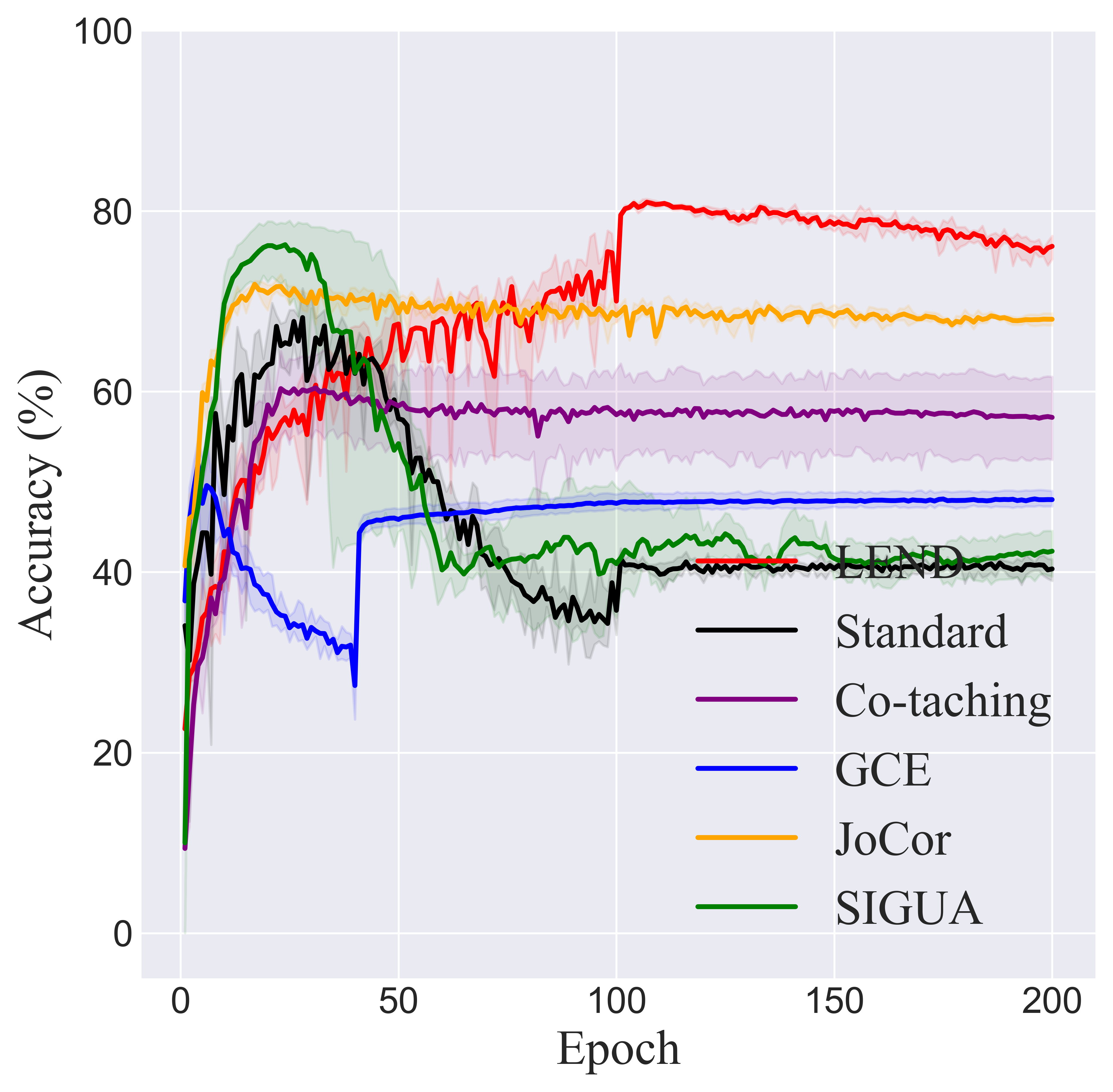}}
	\end{minipage} \vspace{-0.2cm}
	\caption{The curves of test accuracy on \textit{CIFAR-10} dataset under symmetric noise for all compared methods. The colored curves show the mean accuracy of five trials, and the shaded bars denote the standard deviations of the accuracies over five trials. }
	\label{sym}
\end{figure*}

\begin{figure*}[t]
	\begin{minipage}{1\linewidth}
		\centering
		\subfigure[Asymmetric-20\%.]
		{\includegraphics[width=0.3\linewidth]{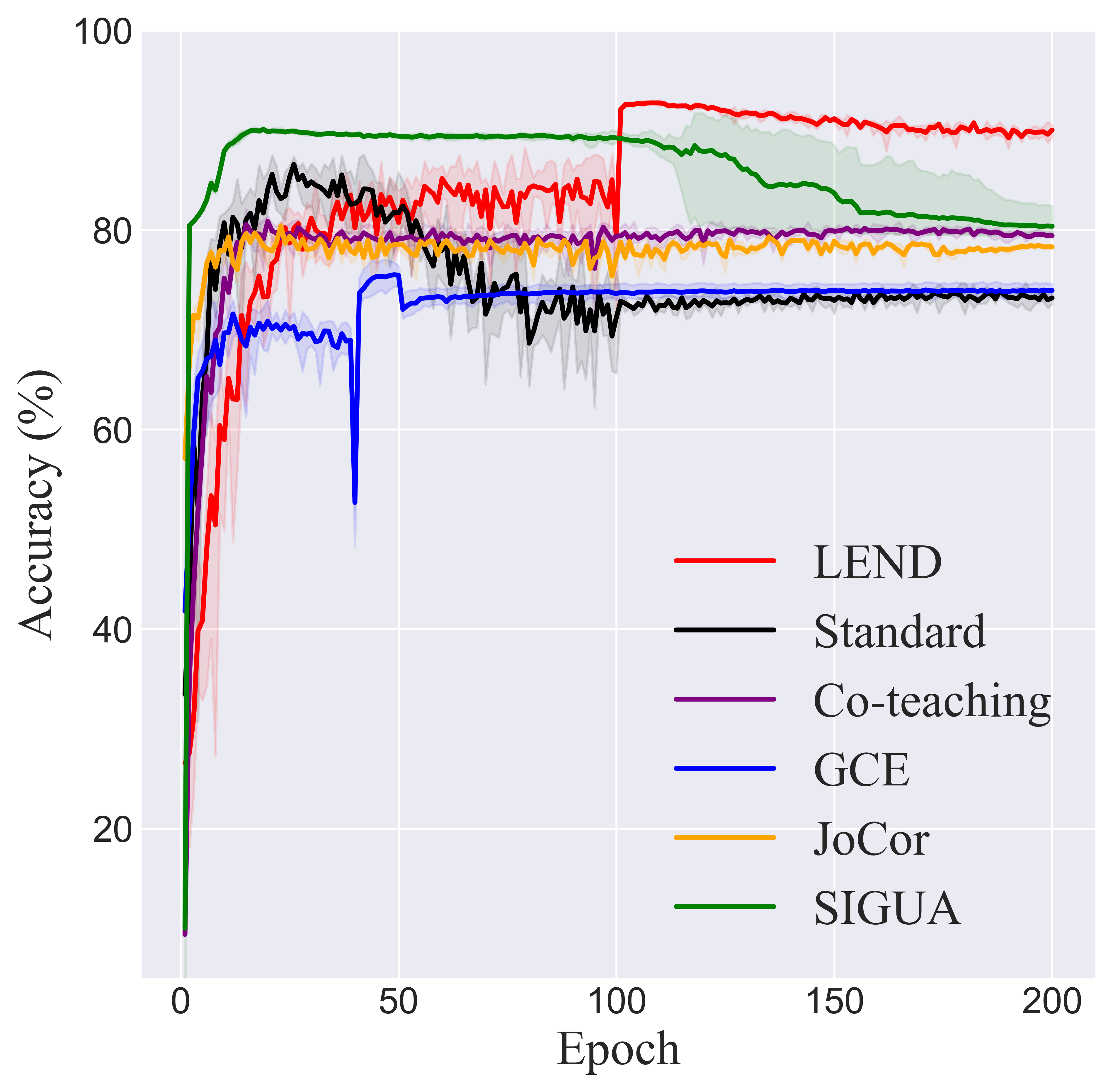}}		
		\subfigure[Asymmetric-30\%.]
		{\includegraphics[width=0.3\linewidth]{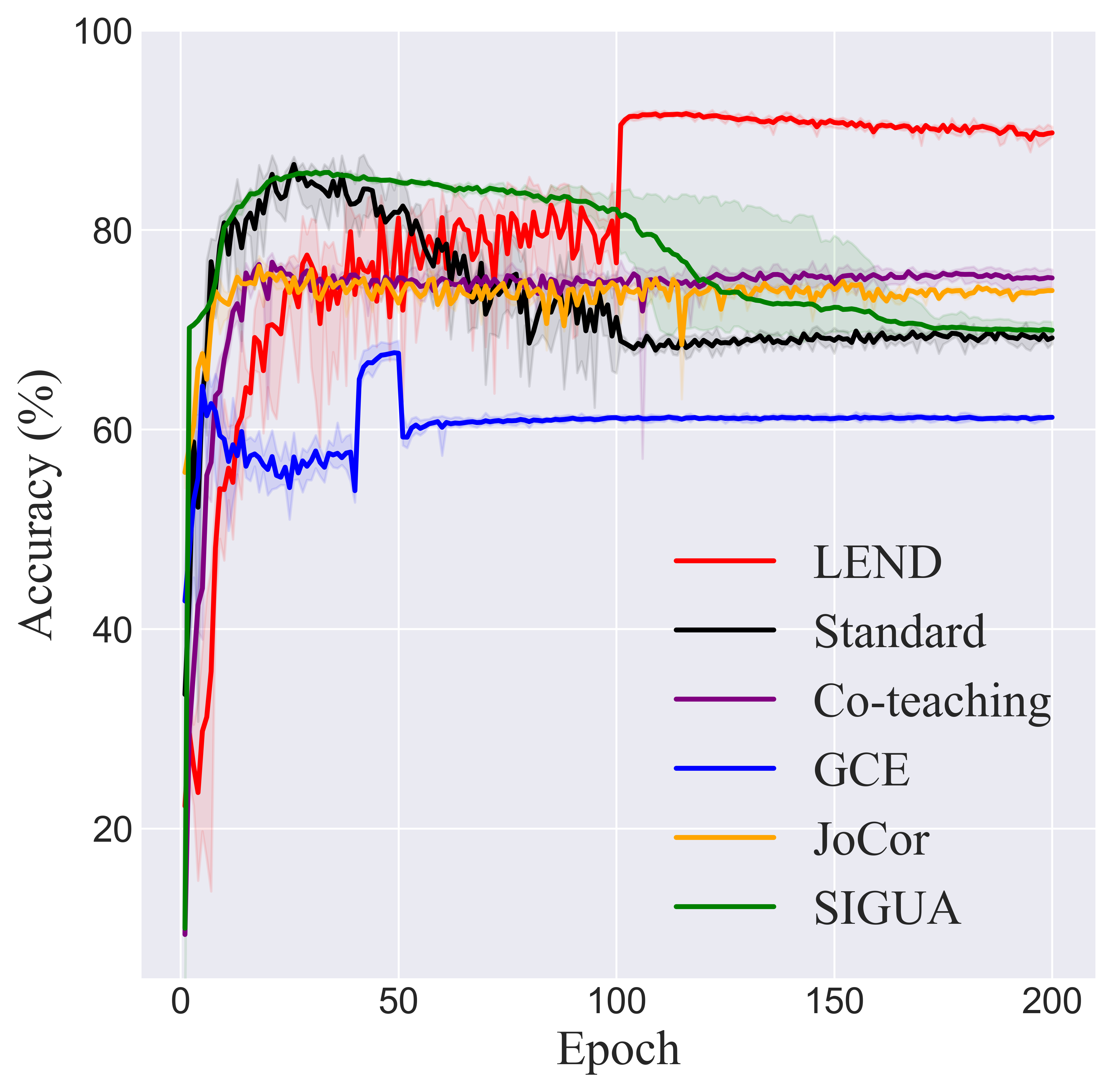}}
		\subfigure[Asymmetric-40\%.]
		{\includegraphics[width=0.3\linewidth]{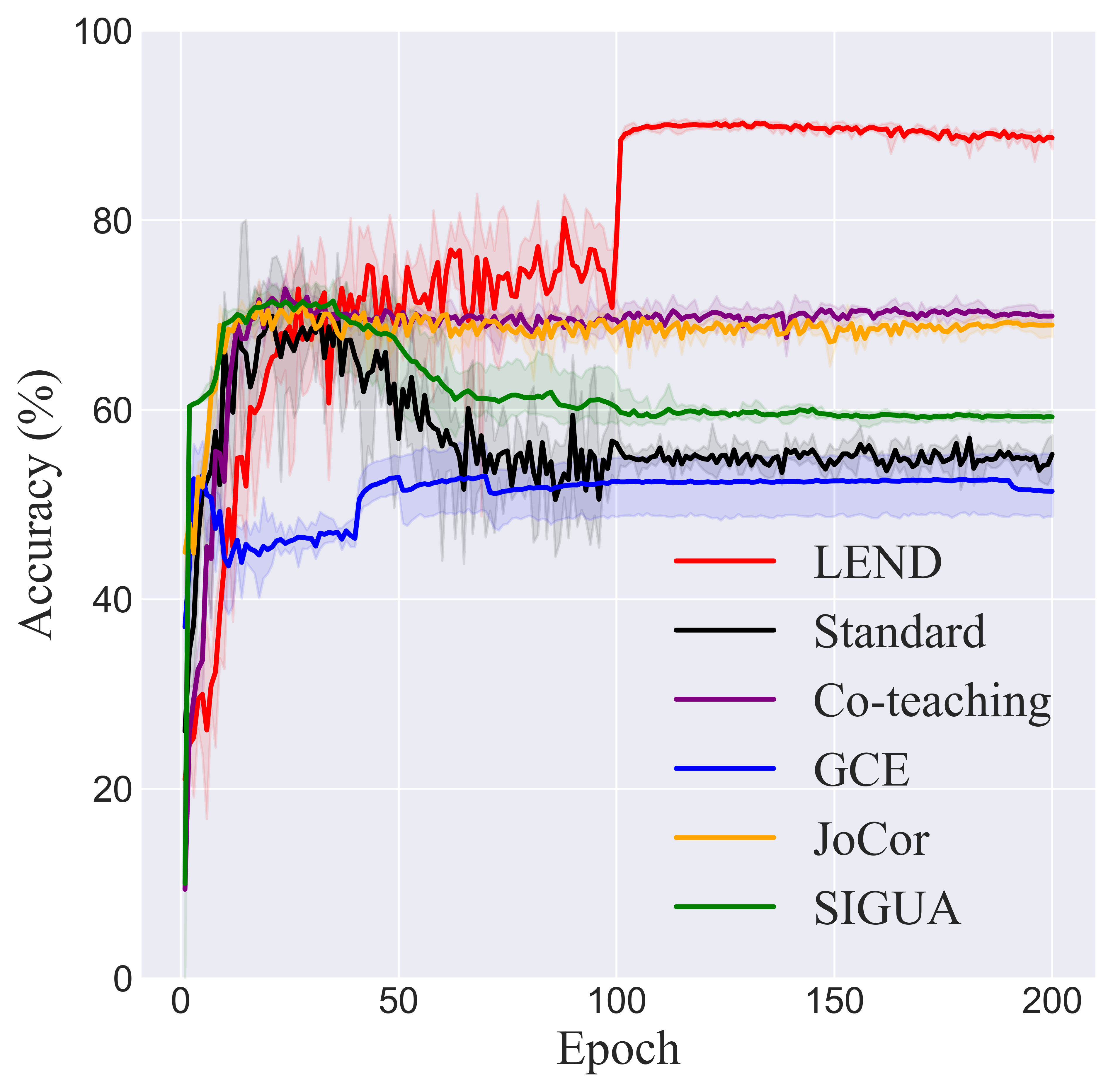}}
	\end{minipage} \vspace{-0.2cm}
	\caption{The curves of test accuracy on \textit{CIFAR-10} dataset under asymmetric noise for all compared methods. The colored curves show the mean accuracy of five trials, and the shaded bars denote the standard deviations of the accuracies over five trials. }
	\label{asym}
\end{figure*}

	\paragraph{Results on CIFAR-10}
	For \textit{CIFAR-10}, we evaluate the proposed method with synthetic label noise,  where symmetric label noise with the noise rates $\rho \in \{0.4,0.5,0.6\}$ and asymmetric label noise with the noise rates $\rho \in \{0.2,0.3,0.4\}$ are included. We run five individual trials for all compared methods under each noise level. Fig.~\ref{sym} and Fig.~\ref{asym} plot the test accuracy \textit{v.s.} the number of epochs regarding the symmetric and asymmetric cases, respectively. In all plots, we can clearly see the memorization effects of DNNs. For example, the test accuracy of ``Standard'' first reaches a very high level since DNN will first fit clean labels. Over the increase of epochs, the deep network will over-fit noisy labels gradually, which decreases its test accuracy accordingly. By contrast, our LEND increases steadily, verifying its effectiveness in alleviating the accuracy decreasing caused by the memorization effect.
	
    In symmetric cases with low noise rate (\textit{e.g.}, Symmetric-$40\%$ and Symmetric-$50\%$), all label noise learning methods work better than``Standard'', which demonstrates their robustness. However, when encountering some difficult cases, such as the symmetric noise with $60\%$ noise rate (Symmetric-$60\%$) and asymmetric cases, SIGUA and GCE suffer from significant accuracy drop, and GCE even performs worse than ``Standard'' in  the challenging asymmetric cases.  By contrast, our LEND  increases steadily over the increase of epochs and finally exceeds all compared methods with a large margin for all label noise cases, verifying the effectiveness of our method in dealing with both symmetric and asymmetric label noise.

\begin{table}[t]
	    \centering
	    \caption{Comparison with the representative methods on \textit{CIFAR-100} dataset under symmetric label noise. We report the best accuracy ($\%$) and the averaged test accuracy ($\%$) over the last ten epochs. The highest records are marked in \textbf{bold}.}
	    \setlength{\tabcolsep}{4mm}
	    \begin{tabular}{ccccc}
	        \toprule
	        \multicolumn{2}{c}{\multirow{2}{*}{Method}} & \multicolumn{3}{c}{Symmetric flipping} \\ 
	        \cline{3-5} 
	        \multicolumn{2}{l}{}     &$40\%$       &$50\%$       &$60\%$      \\ 
	        \midrule
	        
	        \multirow{2}*{Standard}     & \textit{Best}    & 43.75$\pm$0.64    & 36.92$\pm$0.56    & 25.71$\pm$0.63 \\
	                                    & \textit{Last}    & 33.24$\pm$0.41    & 21.80$\pm$0.38    & 15.33$\pm$0.32 \\
	        \specialrule{0em}{2pt}{1pt}
	        
	        \multirow{2}*{Co-teaching}  & \textit{Best}    & 40.53$\pm$0.49    & 34.98$\pm$0.36    & 28.92$\pm$0.25 \\
	                                    & \textit{Last}    & 38.40$\pm$0.45    & 30.45$\pm$0.21    & 24.79$\pm$0.11 \\
	        \specialrule{0em}{2pt}{1pt}
	        
	        \multirow{2}*{GCE}          & \textit{Best}    & 43.68$\pm$0.04    & 40.17$\pm$0.26    & 40.34$\pm$0.37 \\
	                                    & \textit{Last}    & 43.48$\pm$0.05    & 36.12$\pm$0.12    & 33.39$\pm$0.22 \\
	        \specialrule{0em}{2pt}{1pt}
	        
	        \multirow{2}*{JoCor}        & \textit{Best}    & 45.89$\pm$0.80    & 40.86$\pm$0.62    & 38.25$\pm$0.52 \\
	                                    & \textit{Last}    & 40.46$\pm$0.69    & 34.52$\pm$0.54    & 31.73$\pm$0.43 \\
	        \specialrule{0em}{2pt}{1pt}
	        
	        \multirow{2}*{SIGUA}        & \textit{Best}    & 37.12$\pm$0.35    & 32.78$\pm$0.33    & 30.13$\pm$0.25 \\
	                                    & \textit{Last}    & 36.20$\pm$0.25    & 32.46$\pm$0.14    & 29.66$\pm$0.12 \\
	        \specialrule{0em}{2pt}{1pt}
	        
	        \multirow{2}*{LEND}         & \textit{Best}    & \textbf{52.71$\pm$0.21}    & \textbf{44.37$\pm$0.26}    & \textbf{42.63$\pm$0.18} \\
	                                    & \textit{Last}    & \textbf{48.73$\pm$0.17}    & \textbf{40.37$\pm$0.14}    & \textbf{38.72$\pm$0.12} \\
	        \bottomrule
	    \end{tabular}
        \label{sym100}
	\end{table}

		\begin{table}[t]
	    \centering
	    \caption{Comparison with the representative methods on \textit{CIFAR-100} dataset under asymmetric label noise. We report the best accuracy ($\%$) and the averaged test accuracy ($\%$) over the last ten epochs. The highest records are marked in \textbf{bold}.}
	    \setlength{\tabcolsep}{4mm}
	    \begin{tabular}{ccccc}
	        \toprule
	        \multicolumn{2}{c}{\multirow{2}{*}{Method}} & \multicolumn{3}{c}{Asymmetric flipping} \\ 
	        \cline{3-5} 
	        \multicolumn{2}{l}{}     &$20\%$       &$30\%$       &$40\%$      \\ 
	        \midrule
	        
	        \multirow{2}*{Standard}     & \textit{Best}    & 42.97$\pm$0.26    & 35.53$\pm$0.07    & 30.89$\pm$0.21 \\
	                                    & \textit{Last}    & 41.52$\pm$0.17    & 34.29$\pm$0.17    & 29.43$\pm$0.12 \\
	        \specialrule{0em}{2pt}{1pt}
	        
	        \multirow{2}*{Co-teaching}  & \textit{Best}    & 46.22$\pm$0.51    & 40.08$\pm$0.29    & 32.64$\pm$0.21 \\
	                                    & \textit{Last}    & 46.22$\pm$0.25    & 39.59$\pm$0.61    & 31.49$\pm$0.19 \\
	        \specialrule{0em}{2pt}{1pt}
	        
	        \multirow{2}*{GCE}          & \textit{Best}    & 50.71$\pm$0.57    & 48.22$\pm$0.67    & 45.36$\pm$0.55 \\
	                                    & \textit{Last}    & 50.65$\pm$0.54    & 42.31$\pm$0.15    & 36.53$\pm$0.15 \\
	        \specialrule{0em}{2pt}{1pt}
	        
	        \multirow{2}*{JoCor}        & \textit{Best}    & 46.40$\pm$0.42    & 40.34$\pm$0.47    & 34.07$\pm$0.15 \\
	                                    & \textit{Last}    & 43.60$\pm$0.14    & 38.01$\pm$0.22    & 31.57$\pm$0.08 \\
	        \specialrule{0em}{2pt}{1pt}
	        
	        \multirow{2}*{SIGUA}        & \textit{Best}    & 35.73$\pm$0.30    & 32.34$\pm$0.17    & 28.06$\pm$0.24 \\
	                                    & \textit{Last}    & 34.82$\pm$0.18    & 31.13$\pm$0.10    & 27.97$\pm$0.15 \\
	        \specialrule{0em}{2pt}{1pt}
	        
	        \multirow{2}*{LEND}         & \textit{Best}    & \textbf{61.95$\pm$0.07}    & \textbf{54.32$\pm$0.17}    & \textbf{49.62$\pm$0.21} \\
	                                    & \textit{Last}    & \textbf{54.32$\pm$0.18}    & \textbf{45.42$\pm$0.18}    & \textbf{39.61$\pm$0.17} \\
	        \bottomrule
	    \end{tabular}
        \label{asym100}
	\end{table}
	
	\paragraph{Results on CIFAR-100}
	Similar to the experimental settings on \textit{CIFAR-10}, symmetric label noise with the noise rates $\rho \in \{0.4,0.5,0.6\}$ and asymmetric label noise with the noise rates $\rho \in \{0.2,0.3,0.4\}$ are considered. We run five individual trials for all compared methods under each noise level and report the average test accuracies and the corresponding standard deviations of all compared methods. Table \ref{sym100} and Table \ref{asym100} provide the experimental results of the symmetric cases and asymmetric cases, respectively. We observe that our proposed LEND is consistently the best method among all compared methods. In particular, our LEND outperforms ``Standard'' (a baseline method without tackling label noise) with a large margin, which demonstrates the significance of our LEND in handling label noise. It is worth noting that Co-teaching and JoCoR are both model prediction-based methods, and the results also demonstrate the stronger robustness of the feature embedding  over model prediction.

\subsection{Experiments on Real-world Dataset}

\begin{wraptable}{r}{0.5\textwidth}
    \vspace{-0.6cm}
    \centering
    \small
    \caption{Comparison with the representative methods on \textit{Animal-10N}. We report the averaged test accuracy ($\%$) over the last ten epochs. The best record is marked in \textbf{bold}.}
    \setlength{\tabcolsep}{6mm}
    \begin{tabular}{c| c}
        \toprule
        Method          & Accuracy       \\
        \hline\hline
        Standard        & $66.8\pm0.03$   \\
        \hline
        Co-teaching     & $69.7\pm0.11$  \\
        \hline
        GCE             & $68.7\pm0.04$   \\
        \hline
        JoCor           & $75.7\pm0.12$   \\
        \hline
        SIGUA           & $74.0\pm0.21$   \\
        \hline
        LEND            & \textbf{76.4$\pm$0.18}  \\
        \bottomrule
    \end{tabular}
    \label{animal}
    \vspace{-0.3cm}
\end{wraptable}

Similar to the experimental settings on \textit{CIFAR-10}, we run five individual trials for all compared methods on \textit{Animal-10N}. Note that we do not apply any data augmentation or pre-processing procedures. Table~\ref{animal} shows the average test accuracies and the corresponding standard deviations of all compared methods on \textit{Animal-10N}. We can clearly see that our LEND achieves the highest classification accuracy among all comparators. Therefore, the proposed LEND is effective in handling real-world label noise.

\subsection{Model Behavior Analysis}\label{para_ana}
	In this section, we investigate 
    1) the accuracy variation of our method with related to the increase of the batch size, 
    2) the parametric sensitivity of our approach to $k$, \textit{i.e.}, the number of nearest neighbors chosen during label noise dilution, and 
	3) the sensitivity of our method to the trade-off parameter $\beta$.
	
\begin{figure*}[t]
	\begin{minipage}{1\linewidth}
		\centering
		\subfigure[Analysis of batch size.]
		{\includegraphics[width=0.3\linewidth]{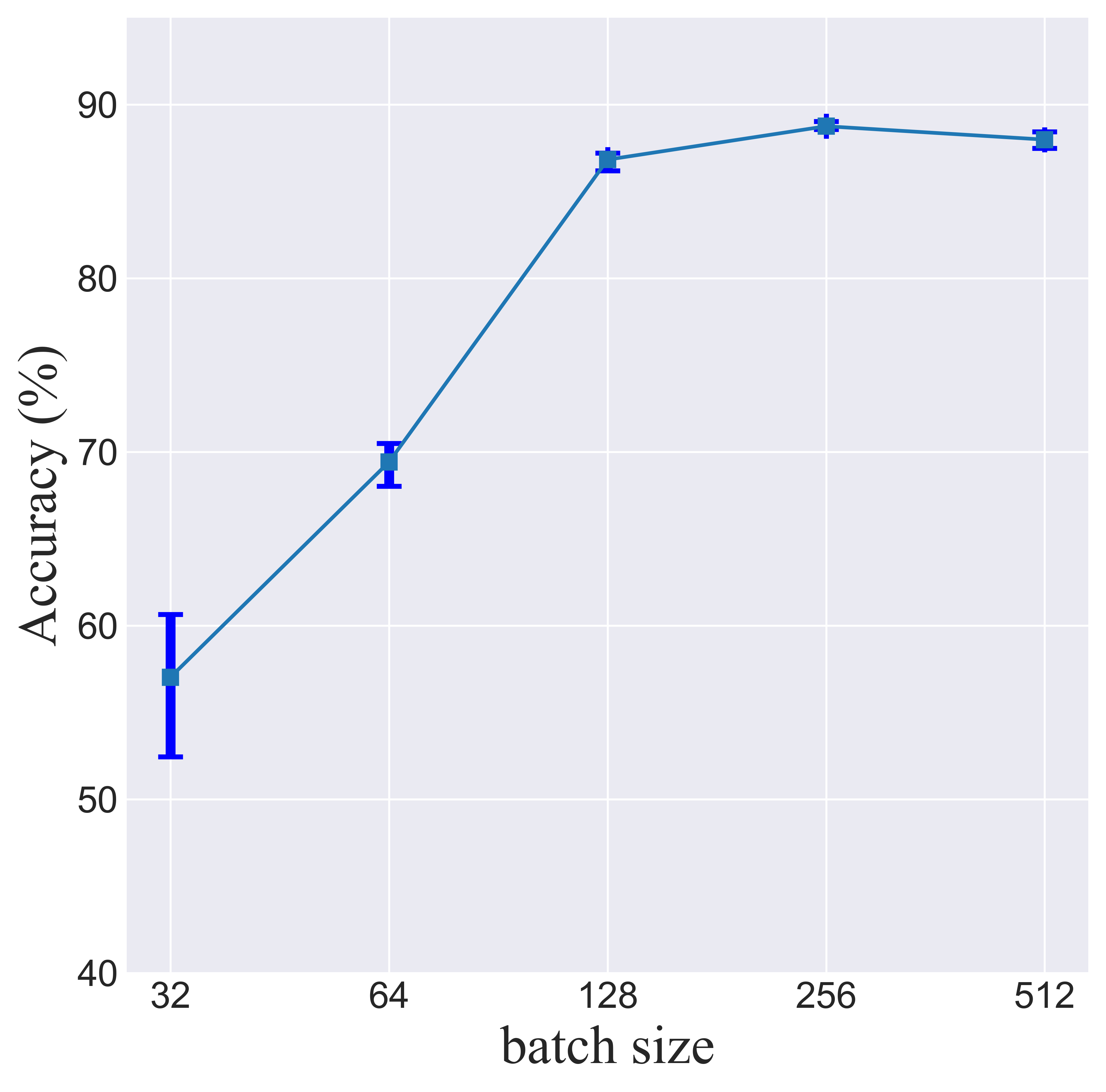}}
		\subfigure[Analysis of $k$.]
		{\includegraphics[width=0.3\linewidth]{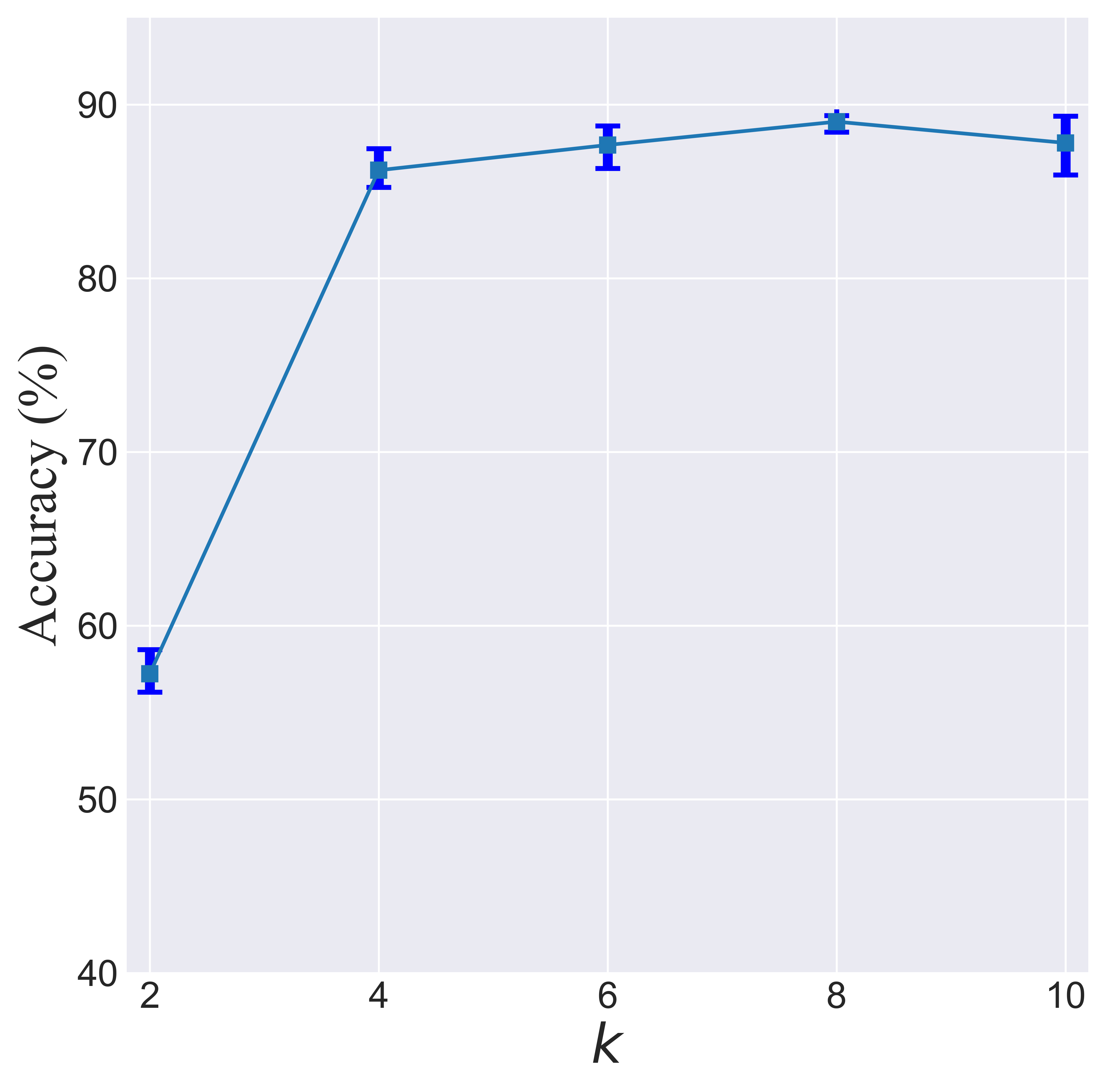}}
		\subfigure[Analysis of $\beta$.]
		{\includegraphics[width=0.3\linewidth]{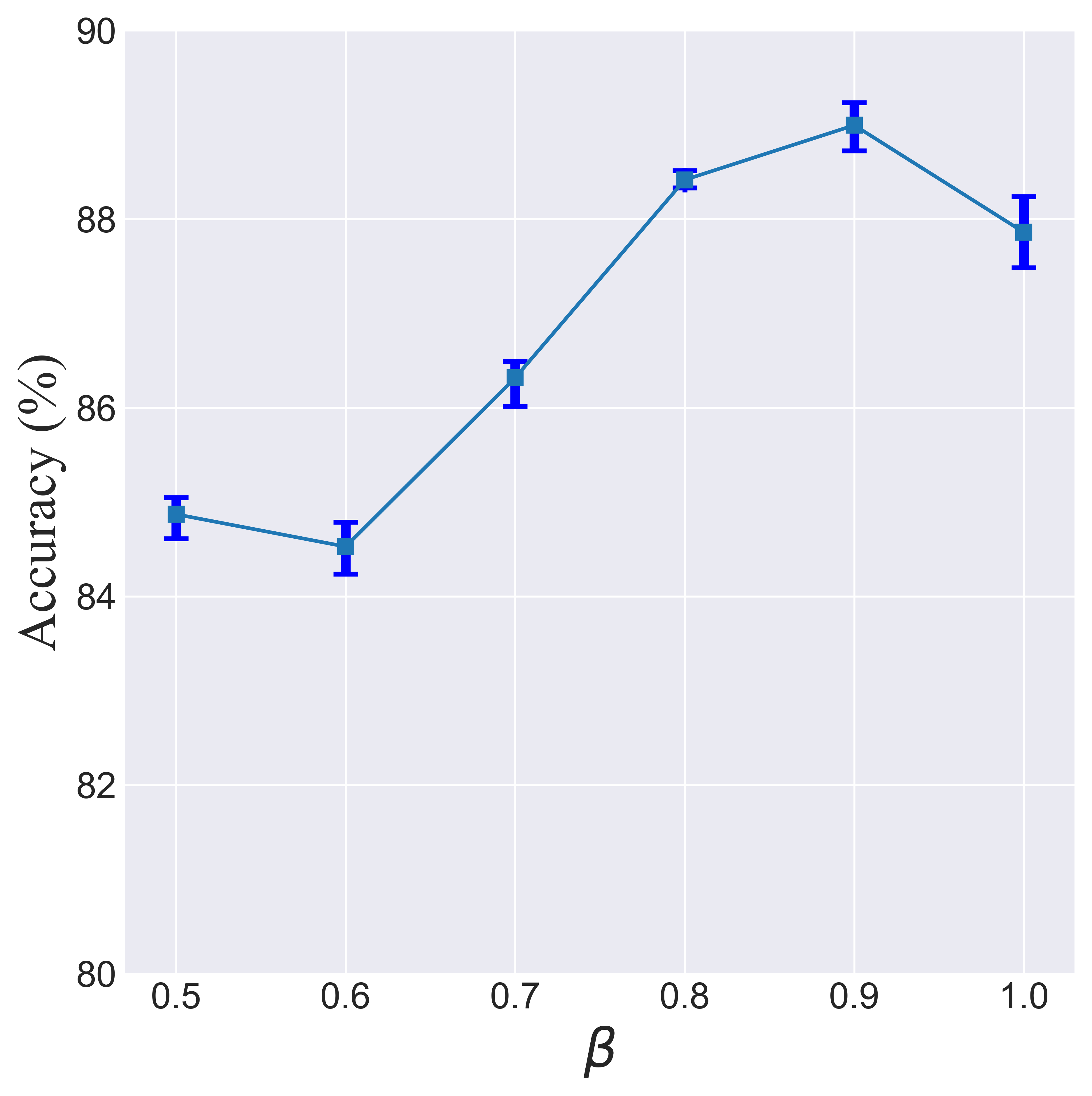}}	
	\end{minipage} \vspace{-0.2cm}
	\caption{Model analysis on \textit{CIFAR-10} dataset under $40\%$ asymmetric noise. (a) presents the analysis of the batch size. (b) shows the model performance when different $k$ is selected. (c) provides the parameter sensitivity of $\beta$.}
	\label{ana}
\end{figure*}

	\textbf{Sensitivity to the batch size.} 
    As mentioned before, we compute a similarity matrix within each mini-batch, which is crucial to the label noise diffusion process. Therefore, it is worth validating the influence of the batch size on the performance of our LEND.
    Fig. \ref{ana} (a) shows the experimental results of our LEND on \textit{CIFAR-10} dataset under $40\%$ asymmetric label noise, with the batch size changing within $\{32, 64, 128, 256, 512\}$. We can see that, with the increase of the batch size, the accuracy of our LEND keeps rising and reaches stable gradually. The reason is that a large batch size can provide more reliable distribution  information and benefits the model performance accordingly. Here, we observe that a batch size of 256 is sufficient to get a satisfactory performance.

	\textbf{Sensitivity to $k$.} Similar to batch size, the number of nearest neighbors, \textit{i.e.}, $k$, is also a crucial factor to label noise dilution. Intuitively, a larger $k$ is preferred, since more nearest neighbors carry more local information.
    Fig. \ref{ana} (b) shows the experimental results of the proposed method on \textit{CIFAR-10} dataset under $40\%$ asymmetric label noise, with $k$ changing within $\{2, 4, 6, 8, 10\}$.  We can see that with the increase of $k$,  the accuracy of our LEND keeps rising and reaches the highest record when $k=8$. However, our LEND shows a slight drop and significant oscillations when $k$ keeps increasing. The reason is that more nearest neighbors will inevitably introduce more noisy examples that do not belong to the class of the examples in this local region. Therefore, we suggest $k=8$ in our experiments.
	
	\textbf{Sensitivity to $\beta$.} In our method, the trade-off parameter $\beta$ plays a key role in tackling the label noise. To provide some guidelines for hyperparameter tuning, we show the curves of validation accuracy during training in Fig. \ref{ana} (c), given $\beta$ within $\{0.5, 0.6, 0.7, 0.8, 0.9, 1.0\}$. We can observe that $\beta=0.9$ leads to satisfactory performance.

	\section{Conclusion}\label{sec:conclusion}
	In this paper, we reveal that the memorization effect induces a good feature embedding in the early stage of learning, and the embedded features show more robustness than the model predictions. Based on this observation, we further propose a simple yet effective feature embedding-based label noise learning method, which can make full use of the memorization effect. To be specific, we propose to dilute the label noise with the help of the strong semantic information retained in the embedded features. As a result, weak semantic information from mislabeled data is overwhelmed by nearby correctly labeled ones, and then the corrected labels are further employed to train a robust classifier. Empirical results on both symmetric and asymmetric label noise cases have demonstrated the effectiveness of the proposed LEND. In the future, it is worthwhile to explore more effective techniques to make full use of the robustness of feature embedding in the early learning stage.
\section*{Acknowledgments}

This work is supported by NSF of China (No. 61973162), NSF of Jiangsu Province (No. BZ2021013), the Fundamental Research Funds for the Central Universities (No. 30920032202 and No. 30921013114), and Science and Technology Innovation 2030 – ``Brain Science and Brain-like Research” Major Project (No. 2021ZD0201402 and No. 2021ZD0201405).

\newpage

\bibliographystyle{plainnat}
\bibliography{sn-bibliography}
\clearpage
	
	\section{Detailed t-SNE Visualizations of Training with Noisy Labels}

	\label{sec:tsne}
\begin{figure*}[h]
		\centering
			\includegraphics[width=6in]{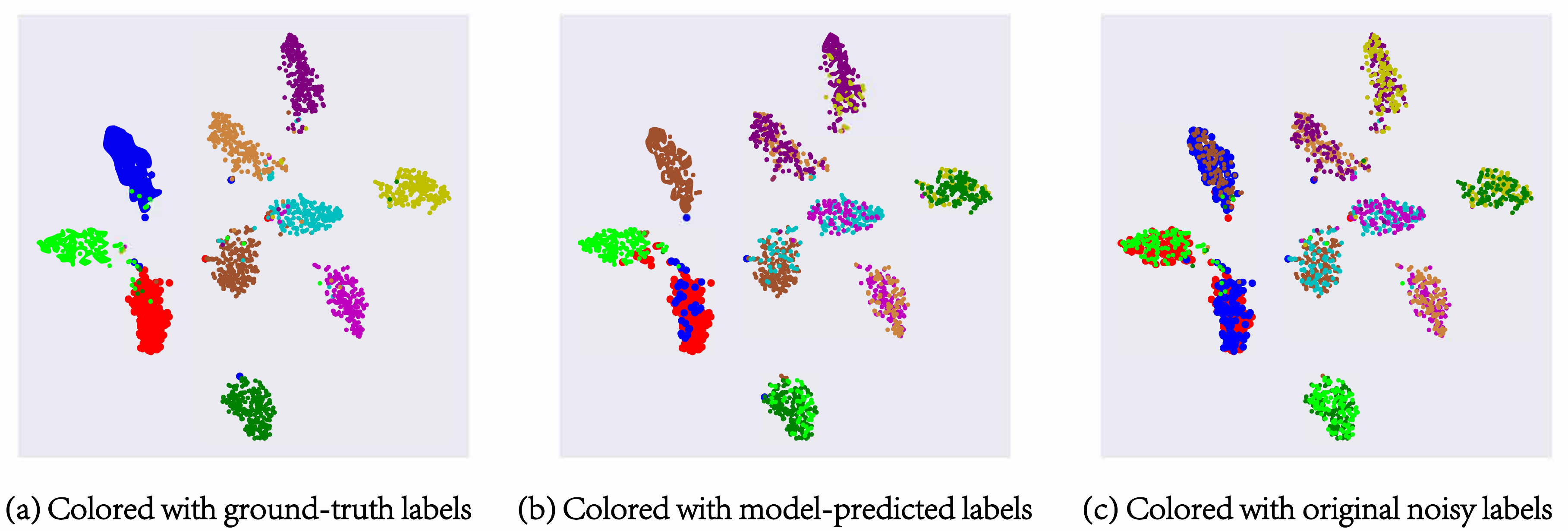}\\
			\caption{The t-SNE visualization of the training data under 50 epochs when training under $45\%$ asymmetric label noise on \textit{CIFAR-10} dataset.} 
			\label{fig:tsne50}
	\end{figure*}

\begin{figure*}[h]
		\centering
			\includegraphics[width=6in]{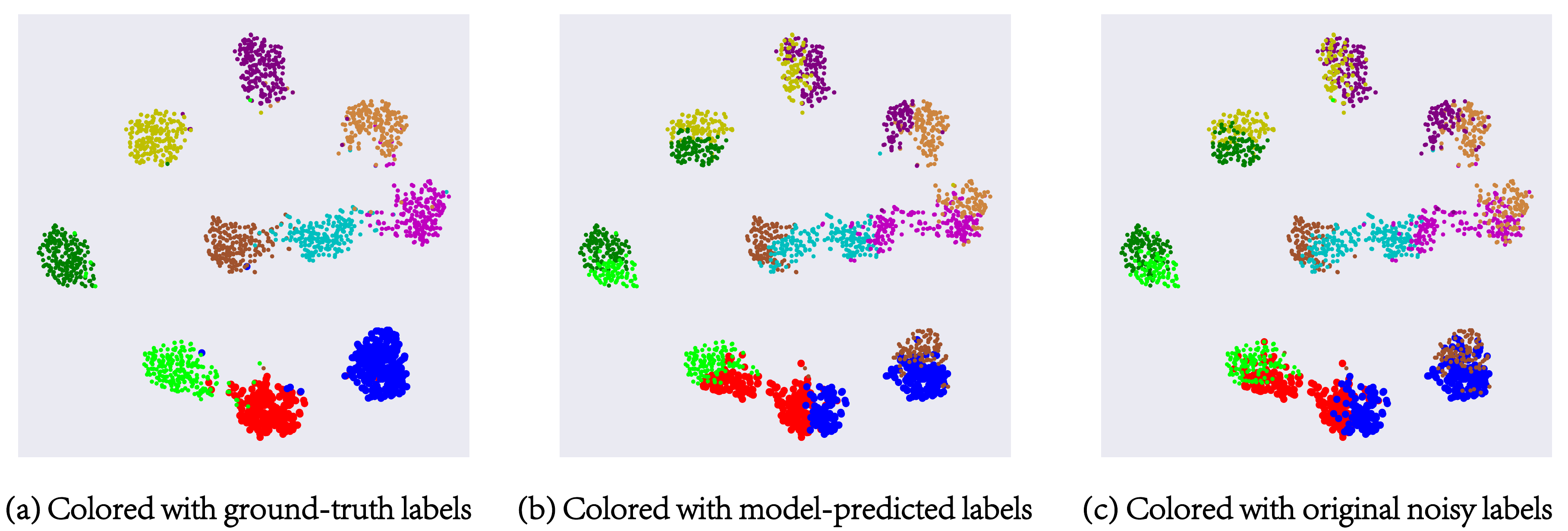}\\
			\caption{The t-SNE visualization of the training data under 100 epochs when training under $45\%$ asymmetric label noise on \textit{CIFAR-10} dataset. }
			\label{fig:tsne100}
	\end{figure*}

\begin{figure*}[t]
		\centering
			\includegraphics[width=6in]{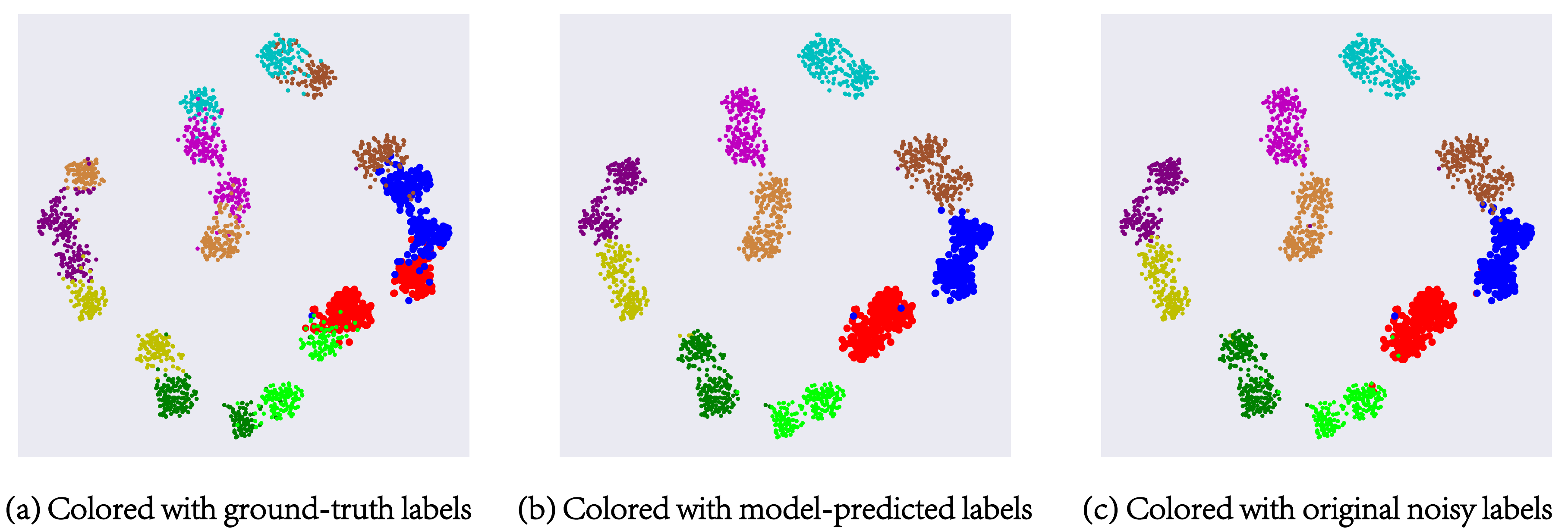}\\
			\caption{The t-SNE visualization of the training data under 150 epochs when training under $45\%$ asymmetric label noise on \textit{CIFAR-10} dataset. }
			\label{fig:tsne150}
	\end{figure*}

\begin{figure*}[t]
		\centering
			\includegraphics[width=6in]{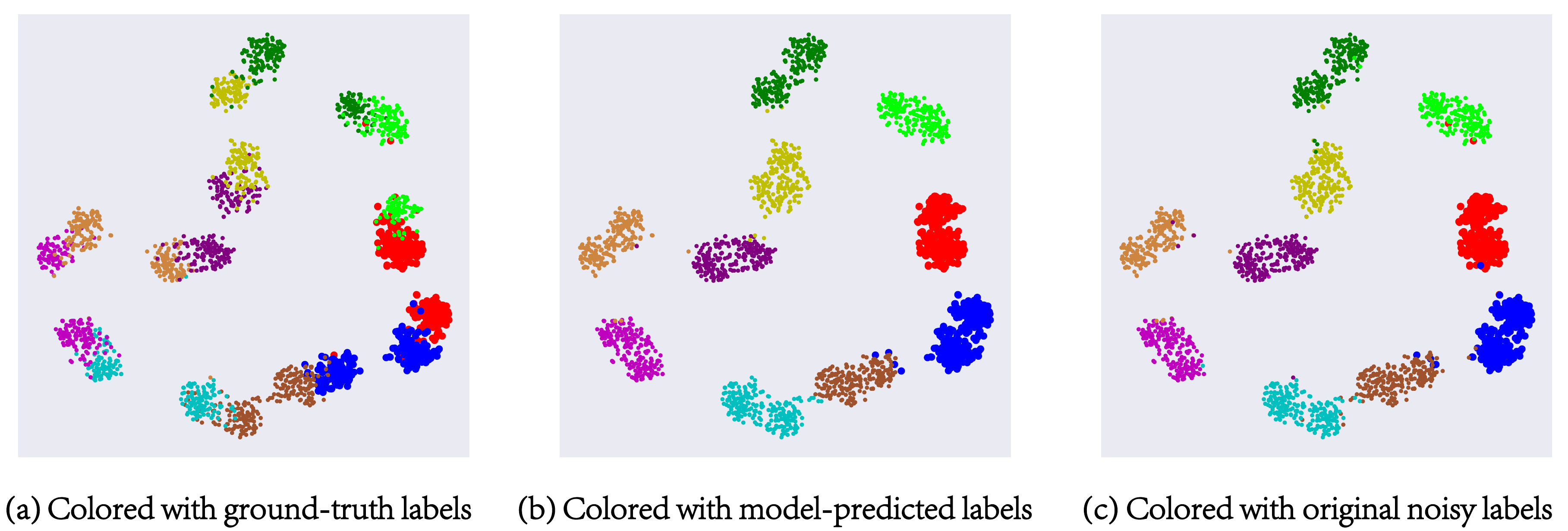}\\
			\caption{The t-SNE visualization of the training data under 200 epochs when training under $45\%$ asymmetric label noise on \textit{CIFAR-10} dataset. }
			\label{fig:tsne200}
	\end{figure*}

	Here, we provide more detailed visualizations of the features of training data on \textit{CIFAR-10} under $45\%$ asymmetric label noise.  To be specific, we plot the t-SNE visualizations under $\{50,100,150,200\}$ epochs of training on Fig. $\{\ref{fig:tsne50},\ref{fig:tsne100},\ref{fig:tsne150},\ref{fig:tsne200}\}$, respectively. ResNet-18 is directly used to fit the noisy data, and the embedded features of the last hidden layer of the neural network are visualized by the t-SNE method. The ground-truth labels, model-predicted labels, and original noisy labels are colored on subfigures (a), (b), and (c), respectively.
	
	We have the following observations:
	\begin{itemize}
	    \item [(1)] When trained with noisy data, DNNs will gradually fit the noisy labels and lead to a good embedding\footnote{A good embedding means that examples with the same labels are pulled together and examples with different labels are pushed away.} with respect to noisy labels at last (see Fig.~\ref{fig:tsne200}).
	    \item [(2)] In the early learning stage, the model predictions are able to recover the distribution of the ground-truth labels to some extent, but still make error-prone predictions (see Fig. \ref{fig:tsne} and ~\ref{fig:tsne50}).
	    \item [(3)] In the early learning stage, the embedded features induce a good embedding with respect to the ground-truth labels (see Fig. \ref{fig:tsne}, ~\ref{fig:tsne50}, and \ref{fig:tsne100}).
	\end{itemize}

	~~~~~
	

\end{document}